\documentclass{article}
\usepackage{iclr2024_conference, times, hyperref, url}

\usepackage{amsmath,amsfonts,bm}

\def\eqref#1{equation~\ref{#1}}

\def\1{\bm{1}}

\DeclareMathAlphabet{\mathsfit}{\encodingdefault}{\sfdefault}{m}{sl}
\SetMathAlphabet{\mathsfit}{bold}{\encodingdefault}{\sfdefault}{bx}{n}

\usepackage{enumitem, graphicx, multirow, amsfonts, amsmath, verbatim, algorithm, algorithmicx, algpseudocode, booktabs, array, arydshln, times, epsfig, amssymb, pifont, capt-of, wrapfig, lipsum, makecell, tabulary, etoolbox, subcaption, color, fontawesome}
\newcommand{\tablestyle}[2]{\setlength{\tabcolsep}{#1}\renewcommand{\arraystretch}{#2}\centering\footnotesize}
\newcommand \blfootnote[1]{
    \begingroup
        \renewcommand
        \thefootnote{}\footnote{#1}
        \addtocounter{footnote}{-1}
        \vspace{-1ex}
    \endgroup
}

\title{Guiding Instruction-based Image Editing via Multimodal Large Language Models}
\author{$^\text{\faApple}$~Tsu-Jui Fu$^1$, Wenze Hu$^2$, Xianzhi Du$^2$, William Yang Wang$^1$, Yinfei Yang$^2$, Zhe Gan$^2$\\$^1$UC Santa Barbara, $^2$Apple}

\iclrfinalcopy

\begin{document}
\maketitle

\begin{figure}[h]
\centering
    \includegraphics[width=\linewidth]{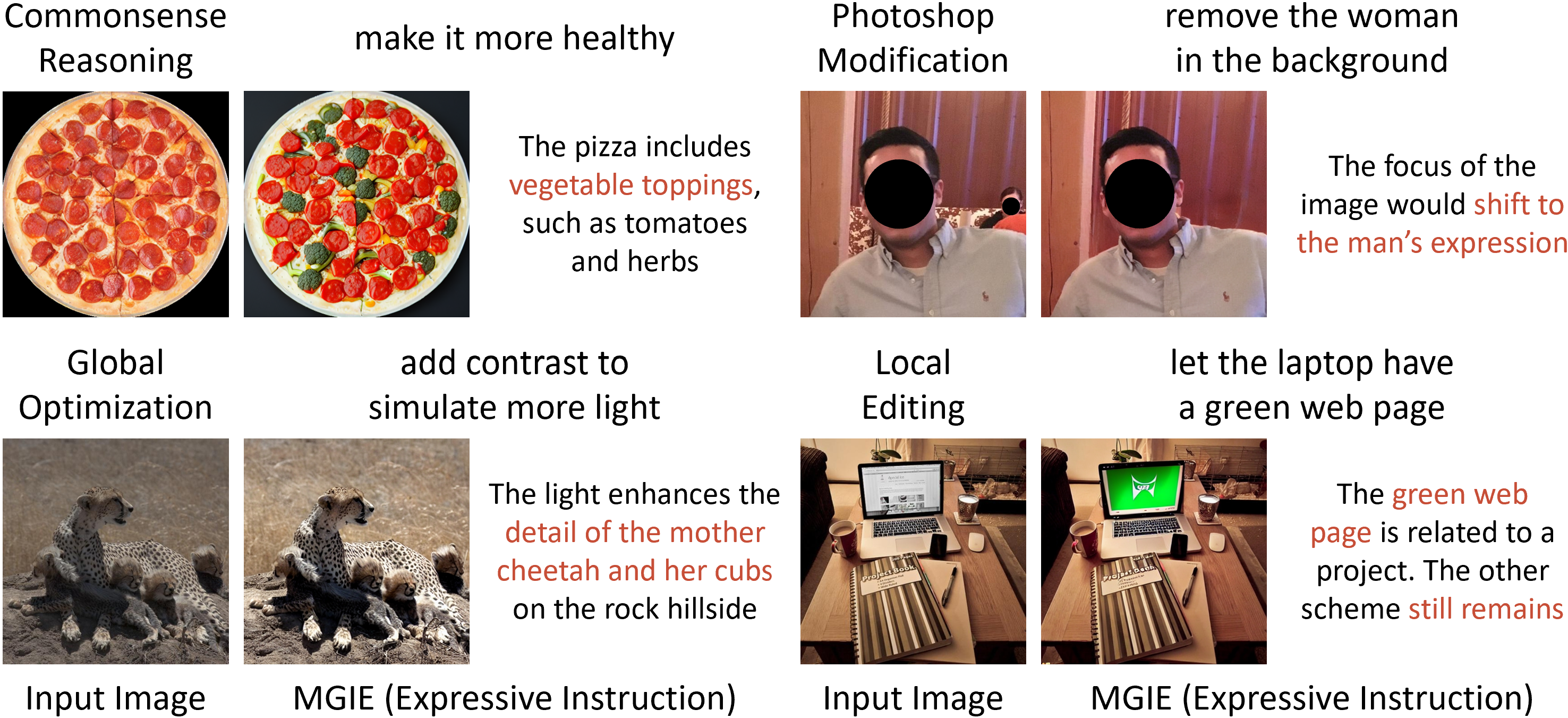}
    \vspace{-3ex}
    \caption{We introduce MLLM-Guided Image Editing (MGIE) to improve instruction-based image editing for various editing aspects. The top is the input instruction, and the right is the jointly derived expressive instruction by MGIE.}
    \label{fig:teaser}
\end{figure}

\begin{abstract}
Instruction-based image editing improves the controllability and flexibility of image manipulation via natural commands without elaborate descriptions or regional masks. However, human instructions are sometimes too brief for current methods to capture and follow. Multimodal large language models (MLLMs) show promising capabilities in cross-modal understanding and visual-aware response generation via LMs. We investigate how MLLMs facilitate edit instructions and present MLLM-Guided Image Editing (MGIE). MGIE learns to derive expressive instructions and provides explicit guidance. The editing model jointly captures this visual imagination and performs manipulation through end-to-end training. We evaluate various aspects of Photoshop-style modification, global photo optimization, and local editing. Extensive experimental results demonstrate that expressive instructions are crucial to instruction-based image editing, and our MGIE can lead to a notable improvement in automatic metrics and human evaluation while maintaining competitive inference efficiency.\blfootnote{\faApple~Work done during an internship at Apple. Project website:~\url{https://mllm-ie.github.io}}
\end{abstract}

\section{Introduction}
Visual design tools are widely adopted in various multimedia fields nowadays. Despite considerable demand, they require prior knowledge to operate. To enhance controllability and accessibility, text-guided image editing has obtained popularity in recent studies~\citep{li2020mani-gan,patashnik2021style-clip,crowson2022vq-clip,gal2022nada}. With an attractive ability to model realistic images, diffusion models~\citep{ho2020diffusion} are also adopted in image editing~\citep{kim2022diffusion-clip}. By swapping the latent cross-modal maps, models can perform visual manipulation to reflect the alteration of the input-goal caption~\citep{hertz2023prompt-to-prompt,mokady2023null-text,kawar2023imagic}. They can further edit a specific region by a guided mask~\citep{nichol2022glide,avrahami2022blend-diff}. Instead of relying on elaborate descriptions or regional masks, instruction-based editing~\citep{el-nouby2019geneva,li2020mani-gan,fu2020sscr} allows human commands that directly express how and which aspect of an image to edit. This flexibility also benefits practicality as such guidance is more aligned with human intuition.

Due to the data scarcity of the input-goal-instruction triplet, InsPix2Pix~\citep{brooks2023ins-pix2pix} collects a curated IPr2Pr dataset. The instruction is generated by GPT-3~\citep{brown2020gpt-3}, and the input-goal image pair is synthesized from Prompt-to-Prompt~\citep{hertz2023prompt-to-prompt}. InsPix2Pix then applies a pre-trained CLIP text encoder~\citep{radford2021clip} to lead the diffusion model along with the input image. Although having feasible results, CLIP is trained for static descriptions, which is challenging to capture the essential visual transformation in editing. Furthermore, the instruction is too brief but ambiguous and insufficient to guide toward the intended goal. The deficiency limits the effectiveness of InsPix2Pix in instruction-based image editing.

Large language models (LLMs)~\citep{brown2020gpt-3,touvron2023llama} have shown significant advancement in diverse language tasks, including machine translation, text summarization, and question answering. Learning from large-scale corpora with diverse content, LLMs contain latent visual knowledge and creativity, which can assist various vision-and-language tasks~\citep{wu2023visual-chatgpt,feng2023layout-gpt,chakrabarty2023spy}. Upon LLMs, multimodal large language models (MLLMs) can treat images as input naturally and provide visual-aware responses to serve as multimodal assistants ~\citep{zhang2023llama-adapter,liu2023llava,zhu2023mini-gpt-4,koh2023gill}.

\begin{figure}[t]
\centering
    \includegraphics[width=\linewidth]{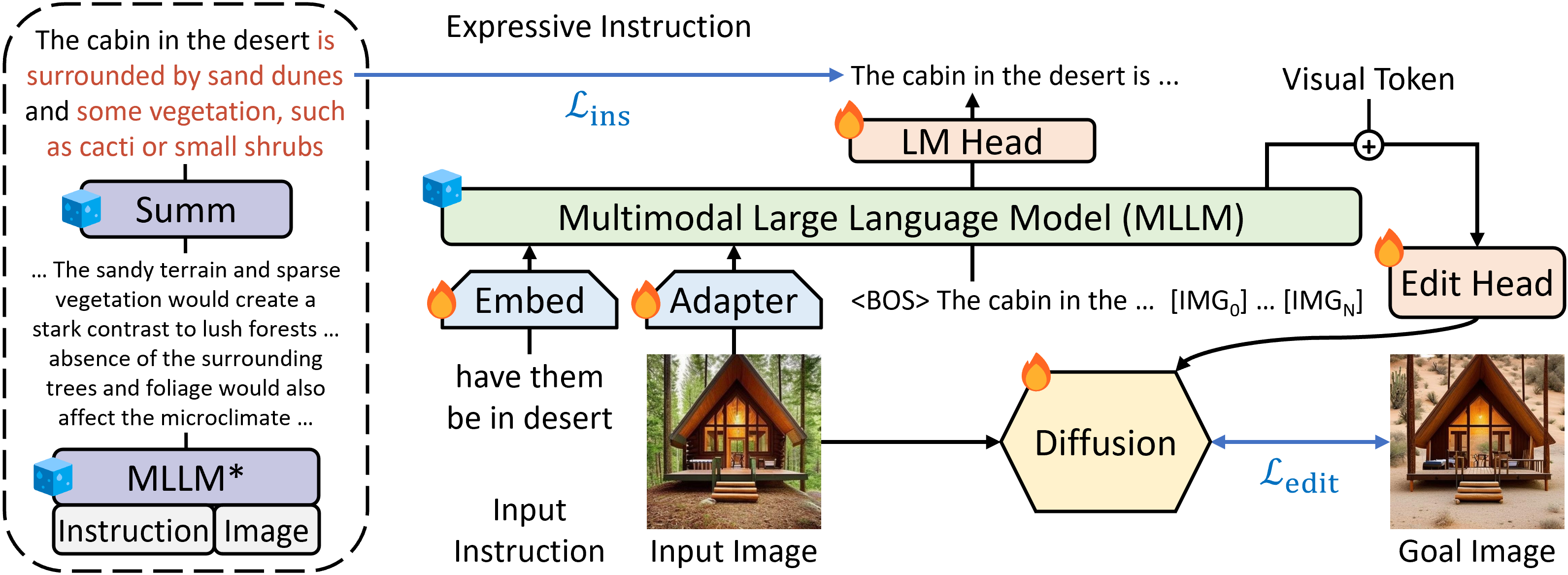}
    \vspace{-3ex}
    \caption[overview]{Overview of MLLM-Guided Image Editing (\textbf{MGIE}), which leverages MLLMs to enhance instruction-based image editing. MGIE learns to derive concise expressive instructions and provides explicit visual-related guidance for the intended goal. The diffusion model jointly trains and achieves image editing with the latent imagination through the edit head in an end-to-end manner.\includegraphics[height=8pt]{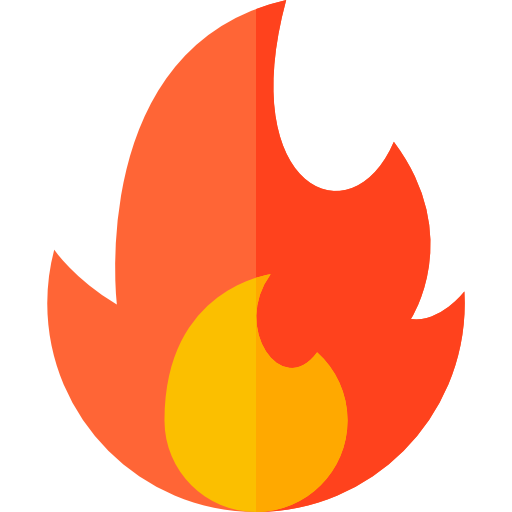} and \includegraphics[height=8pt]{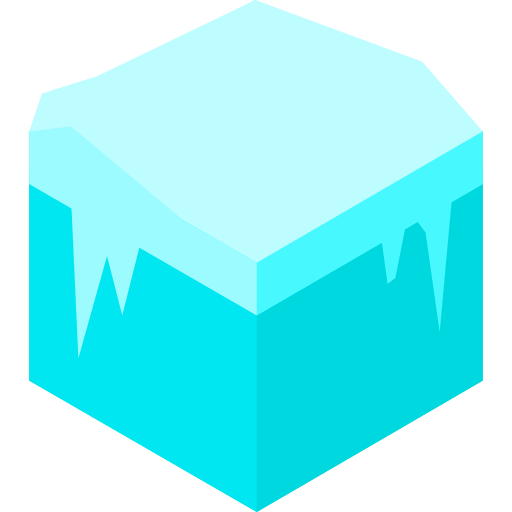} show the module is trainable and frozen\footnotemark, respectively.}
    \label{fig:mgie}
    \vspace{-3ex}
\end{figure}

Inspired by MLLMs, we incorporate them to deal with the insufficient guidance issue of instructions and introduce MLLM-Guided Image Editing (MGIE). As demonstrated in Fig.~\ref{fig:mgie}, MGIE consists of an MLLM and a diffusion model. The MLLM learns to derive concise expressive instructions and offers explicit visual-related guidance. The diffusion model is jointly updated and performs image editing with the latent imagination of the intended goal via end-to-end training. In this way, MGIE benefits from the inherent visual derivation and addresses ambiguous human commands to achieve reasonable editing. For the example in Fig.~\ref{fig:teaser}, it is difficult to capture what ``\textit{healthy}'' means without additional context. Our MGIE can precisely connect ``\textit{vegetable toppings}'' with the pizza and lead to the related editing as human expectation.

To learn instruction-based image editing, we adopt IPr2Pr as our pre-training dataset. We consider different editing aspects in EVR~\citep{tan2019evr}, GIER~\citep{shi2020gier}, MA5k~\citep{shi2022ma5k}, and MagicBrush~\citep{zhang2023magic-brush}. MGIE performs Photoshop-style modification, global photo optimization, and local object alteration. All should be guided by human instructions. Experimental results indicate that our MGIE significantly strengthens instruction-based image editing with reasonable expressive instructions in automatic metrics and human evaluation, and visual-aware guidance is crucial to this improvement. In summary, our contributions are three-fold:
\begin{itemize}[topsep=0pt, noitemsep, leftmargin=*]
    \item We introduce MLLM-Guided Image Editing (MGIE), which jointly learns the MLLM and editing model with visual-aware expressive instructions to provide explicit guidance.
    \item We conduct comprehensive studies from various editing aspects, including Photoshop-style modification, global photo optimization, and local editing, along with qualitative comparisons.
    \item Extensive experiments demonstrate that visual-aware expressive instructions are crucial for image editing, and our MGIE effectively enhances editing performance.
\end{itemize}

\section{Related Work}
\paragraph{Instruction-based Image Editing.}
Text-guided image editing can significantly improve the controllability and accessibility of visual manipulation by following human commands. Previous works built upon the GAN frameworks~\citep{goodfellow2015gan,reed2016t2i} to alter images but are limited to unrealistic synthesis or specific domains~\citep{nam2018ta-gan,li2020mani-gan,el-nouby2019geneva,fu2020sscr,fu2022ldast}. With promising large-scale training, diffusion models~\citep{ho2020diffusion,ramesh2022dalle2,saharia2022imagen,rombach2022sd} can accomplish image transformation via controlling the cross-modal attention maps for the global caption~\citep{meng2022sde-edit,hertz2023prompt-to-prompt,kawar2023imagic,gu2023photoswap}. Local image editing allows fine-grained manipulation by inpainting target regions with user-provided~\citep{nichol2022glide,avrahami2022blend-diff,wang2023edit-bench} or predicted masks~\citep{bar-tal2022text2live,couairon2023diff-edit} while preserving the remaining areas. Different from them, instruction-based image editing accepts straight commands, such as ``\textit{add fireworks to the sky}'', which is not restricted to elaborate descriptions or regional masks. Recent methods learn from synthetic input-goal-instruction triples~\citep{brooks2023ins-pix2pix} and with additional human feedback~\citep{zhang2023hive} to follow editing instructions. However, the frozen CLIP text encoder is pre-trained for static descriptions but not the crucial transformation in editing. Moreover, the instructions are sometimes ambiguous and imprecise for the editing goal. In this paper, we learn with multimodal large language models to perceive images along with given prompts for expressive instructions, which provides explicit yet detailed guidance, leading to superior editing performance.

\paragraph{Large Language Models for Vision.}
Large language models (LLMs) have demonstrated impressive capabilities for text generation and generalizability in various tasks~\citep{brown2020gpt-3,chowdhery2022palm,touvron2023llama}. With robust text understanding, previous works adapt LLMs for input prompts and reason downstream vision-and-language tasks~\citep{zhang2023mcot,wu2023visual-chatgpt,lu2023chameleon,yang2023mm-react,chakrabarty2023spy}. They further produce pseudocode instructions or executable programs by LLMs~\citep{huang2022llmap,gupta2023vis-prog,surís2023viper-gpt,feng2023layout-gpt,lian2023lmd}. Through visual feature alignment with instruction tuning, multimodal large language models (MLLMs) can perceive images and provide adequate responses~\citep{li2023blip-2,zhang2023llama-adapter,liu2023llava,zhu2023mini-gpt-4}. Recently, studies also adopt MLLMs for generating chat-related images~\citep{koh2023gill,sun2023emu}. However, they can only produce images from scratch, which are distinct from inputs. Our proposed MGIE is the first to leverage MLLMs and improve image editing with derived expressive instructions.

\footnotetext{We adopt Flan-T5-XXL~\citep{chung2022flan-t5}, which has been specifically fine-tuned for summarization, as our summarization model for the original MLLM (MLLM*).\label{foot:mllm}}

\section{Method}
\subsection{Background: Multimodal Large Language Models (MLLMs)}
Large language models (LLMs) have shown impressive capabilities for natural language generation. Multimodal large language models (MLLMs) empower LLMs to perceive images and provide reasonable responses. Initialized from a pre-trained LLM, the MLLM contains a visual encoder (\textit{e.g.,} CLIP-L~\citep{radford2021clip}) to extract the visual features $f$, and an adapter $\mathcal{W}$ to project $f$ into the language modality. We follow the training of LLaVA~\citep{liu2023llava}, which is summarized as:
\begin{equation}
\begin{split} \label{eq:mllm}
    \mathcal{C} &= \{x_1, x_2, ..., x_l\}, \\
    f &= \text{Enc}_\text{vis}(\mathcal{V}), \\
    x_t &= \text{MLLM}(\{x_1, ...x_{t-1}\}~|~\mathcal{W}(f)),
\end{split}
\end{equation}
where $l$ is the length of the word token in $\mathcal{C}$. $\mathcal{C}$ can be the image caption (Features Alignment) or the multimodal instruction-following data (Instruction Tuning). The MLLM follows the standard auto-regressive training for the next token prediction and then can serve as a visual assistant for various tasks such as visual question answering and complex reasoning. Although the MLLM is capable of visual perception via the above training, its output is still limited to text.

\subsection{MLLM-Guided Image Editing (MGIE)}
As illustrated in Fig.~\ref{fig:mgie}, we propose MLLM-Guided Image Editing (MGIE) to edit an input image $\mathcal{V}$ into a goal image $\mathcal{O}$, by a given instruction $\mathcal{X}$. To handle imprecise instructions, MGIE contains the MLLM and learns to derive explicit yet concise expressive instructions $\mathcal{E}$. To bridge the language and visual modality, we add special \texttt{[IMG]} tokens after $\mathcal{E}$ and adopt the edit head $\mathcal{T}$ to transform them. They serve as the latent visual imagination from the MLLM and guide our diffusion model $\mathcal{F}$ to achieve the intended editing goal. MGIE is then able to comprehend ambiguous commands with visual-related perception for reasonable image editing.

\paragraph{Concise Expressive Instruction.}
From features alignment and instruction tuning, the MLLM can offer visual-related responses with its cross-modal perception. For image editing, we use this prompt ``\textit{what will this image be like if} \texttt{[instruction]}'' as the language input with the image and derive a detailed explanation of the editing command. However, those explanations are always too lengthy and involve redundant descriptions, which even mislead the intention. To obtain succinct narrations, we apply a pre-trained summarizer~\footref{foot:mllm} and make the MLLM learn to generate the summarized outputs. We treat this explicit yet concise guidance as expressive instruction $\mathcal{E}$:
\begin{equation}
\begin{split} \label{eq:ins}
    \mathcal{E} &= \text{Summ}(\text{MLLM*}([\texttt{prompt}, \mathcal{X}]~|~\mathcal{W}(f))) \\
                &= \{w_1, w_2, ..., w_l\}, \\
    w'_t &= \text{MLLM}(\{w_1, ..., w_{t-1}\}~|~\mathcal{W}(f)), \\
    \mathcal{L}_\text{ins} &= \sum\nolimits_{t=1}^{l} \text{CELoss}(w'_t, w_t),
\end{split}
\end{equation}
where we apply the cross-entropy loss (CELoss) to train the MLLM via teacher forcing. $\mathcal{E}$ can provide a more concrete idea than $\mathcal{X}$ such as linking ``\textit{dessert}'' with ``\textit{sand dunes}'' and ``\textit{cacti or small shrubs}'', which mitigates the comprehension gap for reasonable image editing. This strategy further enhances our efficiency. During inference, the trained MGIE straightforwardly derives concise $\mathcal{E}$ instead of rolling out lengthy narrations (22.7 \textit{vs.} 64.5 tokens) and relying on external summarization. MGIE now can acquire a visual imagination of the editing intention but is confined to the language modality. To bridge the gap, we append $N$ visual tokens \texttt{[IMG]} after $\mathcal{E}$, where their word embeddings are trainable, and the MLLM also learns to generate them through its language modeling (LM) head. Inspired by GILL~\citep{koh2023gill}, the visual tokens are treated as visual-related instruction understanding in $\mathcal{E}$ and establish a connection between the language and vision modalities.

\paragraph{Image Editing via Latent Imagination.}
We adopt the edit head $\mathcal{T}$ to transform \texttt{[IMG]} into actual visual guidance. $\mathcal{T}$ is a sequence-to-sequence model, which maps the sequential visual tokens from the MLLM to the semantically meaningful latent $\mathcal{U}=\{u_1, u_2, ..., u_L\}$ as the editing guidance:
\begin{equation}
    u_t = \mathcal{T}(\{u_1, ..., u_{t-1}\}~|~\{e_{\texttt{[IMG]}}+h_{\texttt{[IMG]}}\}),
\end{equation}
where $e$ is the word embedding and $h$ is the hidden state (from the last layer of MLLM before the LM head) of \texttt{[IMG]}. Specifically, the transformation over $e$ can be treated as a general representation in the visual modality, and $h$ is an instance-aware visual imagination for such editing intention. Our $\mathcal{T}$ is similar to GILL and BLIP-2~\citep{li2023blip-2,li2023blip-diff} for extracting visual features.

To guide image editing with the visual imagination $\mathcal{U}$, we consider a latent diffusion model $\mathcal{F}$~\citep{rombach2022sd}, which includes the variational autoencoder (VAE) and addresses denoising diffusion in the latent space. Our goal of $\mathcal{F}$ is to generate the latent goal $o=\text{Enc}_\text{VAE}(\mathcal{O})$ from preserving the latent input $v=\text{Enc}_\text{VAE}(\mathcal{V})$ and following the editing guidance $\{u\}$. The diffusion process keeps adding noises to $o$ as $z_t$, where the noise level is increasing over timesteps $t$. We then learn the UNet $\epsilon_\theta$ to predict the added noise~\citep{ho2020diffusion}. As LDM, we inject the visual imagination into $\epsilon_\theta$ via the cross-attention layer $\text{Attention}(Q, K, V) = \text{softmax}(\frac{QK^T}{\sqrt{\text{dim}}}) \cdot V$ with
\begin{equation}
    Q=W^{(i)}_Q \cdot \varphi_i(z_t), K=W^{(i)}_K \cdot \{u\}, V=W^{(i)}_V \cdot \{u\},
\end{equation}
where $\varphi$ is the flattened operation, $W^{(i)}_Q$, $W^{(i)}_K$, and $W^{(i)}_V$ are learnable attention matrices. Following InsPix2Pix, we also concatenate $v$ with $z_t$. In this way, our $\mathcal{F}$ can condition both $\mathcal{V}$ and $\mathcal{U}$ to perform image editing. We take classifier-free guidance~\citep{ho2021classifier-free}, and the score estimation $s_\theta$ is extrapolated to keep away from the unconditional $\varnothing$, where the editing loss $\mathcal{L}_\text{edit}$ is calculated as:
\begin{equation}
\begin{split} \label{eq:edit}
    s_\theta(z_t, v, \{u\}) &= s_\theta(z_t, \varnothing, \varnothing) \\
    &+ \alpha_\mathcal{V} \cdot (s_\theta(z_t, v, \varnothing) - s_\theta(z_t, \varnothing, \varnothing)) \\
    &+ \alpha_\mathcal{X} \cdot (s_\theta(z_t, v, \{u\}) - s_\theta(z_t, v, \varnothing)), \\
    \mathcal{L}_\text{edit} &= \mathbb{E}_{o, v, \{u\}, \epsilon \sim \mathcal{N}(0, 1), t} \left[ ||\epsilon - \epsilon_\theta(z_t, t, v, \{u\})||^2_2 \right],
\end{split}
\end{equation}
where $\alpha_\mathcal{V}$ and $\alpha_\mathcal{X}$ are the weights of the guidance scale for the image and the instruction. Similar to InsPix2Pix, we randomly make $v=\varnothing$, $\{u\}=\varnothing$, or both $=\varnothing$ for 5\% of data during training. After we have the generated latent $o'$ through the denoising process by $\epsilon_\theta$, we can obtain the editing result $O' = \text{Dec}_\text{VAE}(o')$. During inference, we use $\alpha_\mathcal{V}=1.5$ and $\alpha_\mathcal{X}=7.5$.

\subsection{Learning of MGIE}
\begin{wrapfigure}{r}{0.44\textwidth}
    \vspace{-1.8\intextsep}
    \begin{minipage}{0.44\textwidth}
        \begin{algorithm}[H]
            \begin{algorithmic}[1] \small
                \While{TRAIN\_MGIE}
                    \State $\mathcal{V}$, $\mathcal{X}$, $\mathcal{O}$ $\gets$ input/instruction/goal triple
                    \State $\{w\}$ $\gets$ summarized explanation
                    \State $\{w'\}$ = $\text{MLLM}(\mathcal{V}~|~\mathcal{X})$
                    \State $\mathcal{L}_\text{ins}$ $\gets$ instruction loss \Comment{Eq.~\ref{eq:ins}}
                    \State $\mathcal{U}$ = $\mathcal{T}(\{\texttt{[IMG]}\})$
                    \State $\mathcal{O}'$ = $\mathcal{F}(\mathcal{V}, \mathcal{U})$
                    \State $\mathcal{L}_\text{edit}$ $\gets$ editing loss \Comment{Eq.~\ref{eq:edit}}
                    \State $\mathcal{L}_\text{all}$ $\gets$ overall training loss
                \EndWhile
            \end{algorithmic}
            \caption{MLLM-Guided Image Editing}
            \label{algo:mgie}
        \end{algorithm}
    \end{minipage}
    \vspace{-\intextsep}
\end{wrapfigure}
Algo.~\ref{algo:mgie} presents the learning process of the proposed MGIE. The MLLM learns to derive concise $\mathcal{E}$ via the instruction loss $\mathcal{L}_\text{ins}$. With the latent imagination from \texttt{[IMG]}, $\mathcal{T}$ transforms their modality and guides $\mathcal{F}$ to synthesize the resulting image. The editing loss $\mathcal{L}_\text{edit}$ is applied for diffusion training. Most weights can be frozen (self-attention blocks inside the MLLM), leading to parameter-efficient end-to-end training. Overall optimization of $\mathcal{L}_\text{all}=\mathcal{L}_\text{ins}+0.5\cdot\mathcal{L}_\text{edit}$ can be:
\begin{equation}
    \min_{\text{MLLM}, \mathcal{W}, \mathcal{T}, \mathcal{F}}~~\mathcal{L}_\text{all}\,.
\end{equation}

\begin{table}[t]
\centering \tablestyle{1.2pt}{1.1}
    \begin{tabular}{cccccccccccccccccc}
        \toprule
        \textbf{Method} & ~ & \multicolumn{3}{c}{\textbf{EVR}} & ~ & \multicolumn{3}{c}{\textbf{GIER}} & ~ & \multicolumn{3}{c}{\textbf{MA5k}} & ~ & \multicolumn{4}{c}{\textbf{MagicBrush}} \\
        \cmidrule{3-5} \cmidrule{7-9} \cmidrule{11-13} \cmidrule{15-18} ~ & ~ & L1$\downarrow$ & DINO$\uparrow$ & CVS$\uparrow$ & ~ & L1$\downarrow$ & SSIM$\uparrow$ & CVS$\uparrow$ & ~ & L1$\downarrow$ & SSIM$\uparrow$ & LPIPS$\downarrow$ & ~ & L1$\downarrow$ & DINO$\uparrow$ & CVS$\uparrow$ & CTS$\uparrow$ \\
        \midrule
        InsPix2Pix & ~ & 0.189 & 67.82 & 81.38 & ~ & \underline{0.144} & \underline{57.51} & 86.63 & ~ & 0.176 & 58.92 & 0.359 & ~ & 0.101 & 71.46 & 85.22 & 29.34 \\
        LGIE & ~ & \textbf{0.159} & \underline{69.71} & \textbf{82.04} & ~ & 0.152 & 56.86 & \underline{86.99} & ~ & \underline{0.144} & \underline{64.60} & \underline{0.327} & ~ & \underline{0.084} & \underline{80.90} & \underline{88.87} & \underline{30.10} \\
        MGIE & ~ & \underline{0.163} & \textbf{71.49} & \underline{81.73} & ~ & \textbf{0.135} & \textbf{59.24} & \textbf{88.59} & ~ &  \textbf{0.133} & \textbf{66.25} & \textbf{0.298} & ~ & \textbf{0.082} & \textbf{82.22} & \textbf{91.14} & \textbf{30.40} \\
        \bottomrule
    \end{tabular}
    \vspace{-1.5ex}
    \caption{\textbf{Zero-shot editing results}. All models are only pre-trained on IPr2Pr~\citep{brooks2023ins-pix2pix}.}
    \label{table:zero-shot}
    \vspace{-2ex}
\end{table}

\begin{table}[t]
\centering \tablestyle{1.2pt}{1.1}
    \begin{tabular}{cccccccccccccccccc}
        \toprule
        \textbf{Method} & ~ & \multicolumn{3}{c}{\textbf{EVR}} & ~ & \multicolumn{3}{c}{\textbf{GIER}} & ~ & \multicolumn{3}{c}{\textbf{MA5k}} & ~ & \multicolumn{4}{c}{\textbf{MagicBrush}} \\
        \cmidrule{3-5} \cmidrule{7-9} \cmidrule{11-13} \cmidrule{15-18} ~ & ~ & L1$\downarrow$ & DINO$\uparrow$ & CVS$\uparrow$ & ~ & L1$\downarrow$ & SSIM$\uparrow$ & CVS$\uparrow$ & ~ & L1$\downarrow$ & SSIM$\uparrow$ & LPIPS$\downarrow$ & ~ & L1$\downarrow$ & DINO$\uparrow$ & CVS$\uparrow$ & CTS$\uparrow$ \\
        \midrule
        InsPix2Pix & ~ & 0.166 & 70.79 & 82.76 & ~ & 0.111 & 64.86 & \underline{91.49} & ~ & 0.122 & 67.12 & 0.267 & ~ & 0.063 & 87.99 & 93.83 & 30.93 \\
        LGIE & ~ & \underline{0.147} & \underline{74.71} & \underline{85.06} & ~ & \textbf{0.104} & \underline{65.30} & 90.61 & ~ & \underline{0.094} & \underline{71.47} & \underline{0.246} & ~ & \underline{0.058} & \underline{88.09} & \underline{93.57} & \underline{31.33} \\
        MGIE & ~ & \textbf{0.146} & \textbf{75.65} & \textbf{85.28} & ~ & \underline{0.105} & \textbf{68.68} & \textbf{92.42} & ~ & \textbf{0.082} & \textbf{72.91} & \textbf{0.235} & ~ & \textbf{0.057} & \textbf{90.65} & \textbf{95.28} & \textbf{31.73} \\
        \bottomrule
    \end{tabular}
    \vspace{-1.5ex}
    \caption{\textbf{Fine-tuned editing results}. All models are further fine-tuned and adapted to each dataset.}
    \label{table:fine-tuned}
    \vspace{-3ex}
\end{table}

\section{Experiments}
\subsection{Experimental Setup}
\paragraph{Datasets and Evaluation Metrics.}
We use \textbf{IPr2Pr}~\citep{brooks2023ins-pix2pix} as our pre-training data. It contains 1M CLIP-filtered data, where instructions are extracted by GPT-3~\citep{brown2020gpt-3}, and images are synthesized by Prompt-to-Prompt~\citep{hertz2023prompt-to-prompt}. For a comprehensive evaluation, we consider various editing aspects. \textbf{EVR}~\citep{tan2019evr} collects 5.7K triples from PhotoshopRequest. We treat the standard pixel difference (L1) and visual feature similarity from DINO~\citep{caron2021dino} or the CLIP visual encoder (CVS) between generated images and ground-truth goals as the evaluation metrics. \textbf{GIER}~\citep{shi2020gier} crawls a larger-scale 29.9K triples also from online forums. Since there are more examples about global optimization, we apply L1, CVS, and Structural Similarity Index (SSIM). \textbf{MA5k}~\citep{shi2022ma5k} consists of 24.8K triples and aims at changing the contrast, brightness, or saturation of a whole photo. We leverage L1, SSIM, and Learned Perceptual Image Patch Similarity (LPIPS)~\citep{zhang2018lpips} as the photo difference\footnote{As there is no object alteration in MA5k, feature-based DINO and CVS cannot clearly tell the difference.}. \textbf{MagicBrush}~\citep{zhang2023magic-brush} annotates 10.5K triples. We follow them to use L1, DINO, CVS, and text-visual feature similarity (CTS)~\citep{hessel2021clip-s} between goal captions and resulting images. We treat the same training/validation/testing split as the original settings. Without specific mention, all evaluations are averaged from 5 random seeds in a zero-shot manner, where models are only trained on IPr2Pr.

\begin{figure}[t]
\begin{minipage}{.35\textwidth}
\centering
    \includegraphics[width=\linewidth]{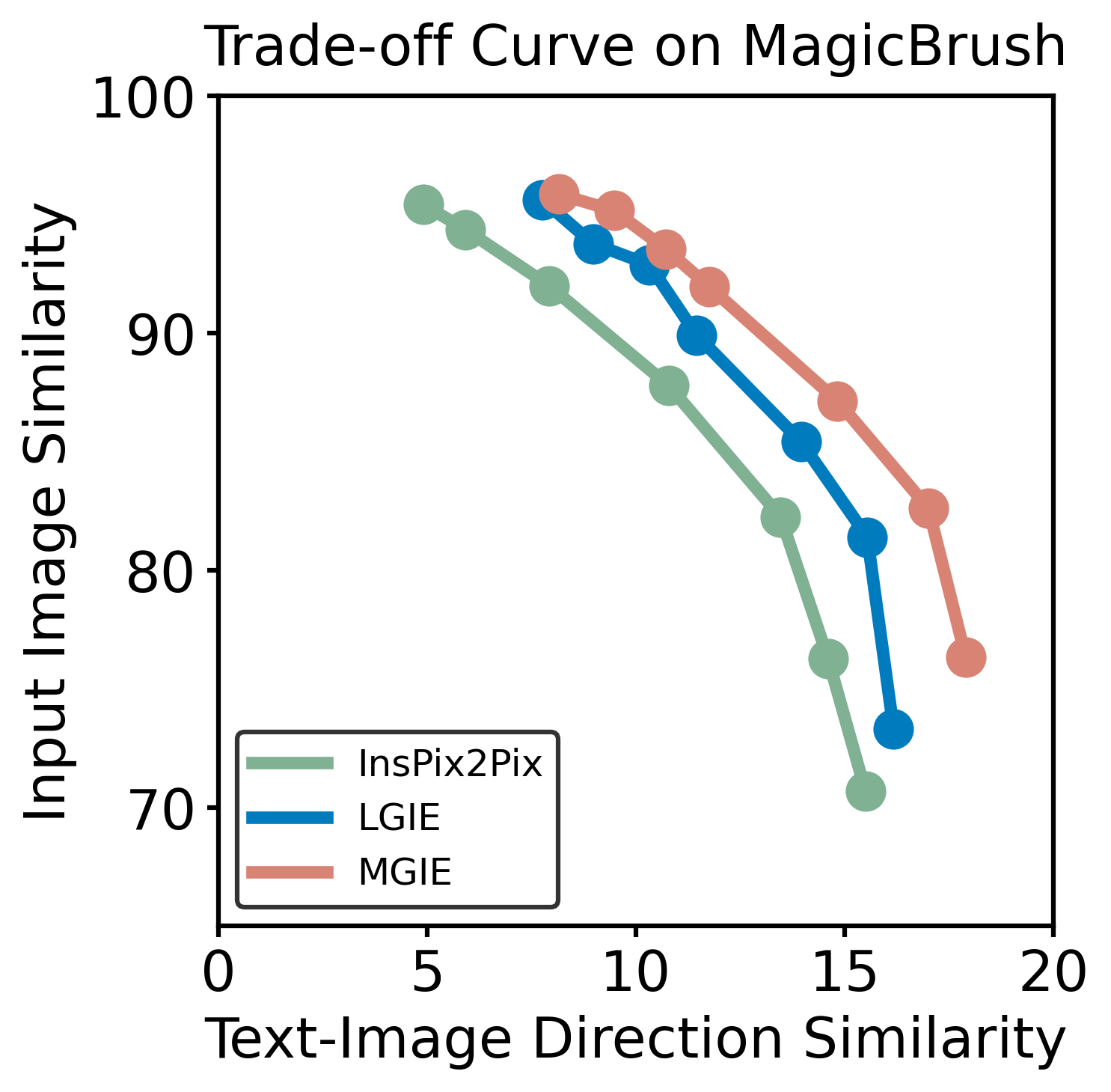}
    \vspace{-3.5ex}
    \captionof{figure}{\textbf{Trade-off curve for image editing}. We set $\alpha_\mathcal{X}$ as 7.5 and vary $\alpha_\mathcal{V}$ in $[1.0, 2.2]$. For both edit (X-axis) and input consistency (Y-axis), higher is better.}
    \label{fig:trade-off}
    \vspace{-3ex}
\end{minipage}~~~~~
\begin{minipage}{.62\textwidth}
\centering \tablestyle{1.2pt}{1.1}
    \begin{tabular}{ccccccccccc}
        \toprule
        \textbf{Arch.} & \textbf{Method} & ~ & \multicolumn{3}{c}{\textbf{MA5k}} & ~ & \multicolumn{4}{c}{\textbf{MagicBrush}} \\
        \cmidrule{4-6} \cmidrule{8-11} ~ & ~ & ~ & L1$\downarrow$ & SSIM$\uparrow$ & LPIPS$\downarrow$ & ~ & L1$\downarrow$ & DINO$\uparrow$ & CVS$\uparrow$ & CTS$\uparrow$ \\
        \midrule
        \multicolumn{2}{c}{InsPix2Pix} & ~ & \underline{0.176} & \textbf{58.92} & \textbf{0.359} & ~ & \textbf{0.101} & \underline{71.46} & \underline{85.22} & \underline{29.34} \\
        \multirow{2}{*}{FZ} & LGIE & ~ & 0.178 & 57.26 & 0.372 & ~ & 0.133 & 67.53 & 82.49 & 28.79 \\
        ~ & MGIE & ~ & \textbf{0.163} & \underline{57.54} & \underline{0.366} & ~ & \underline{0.128} & \textbf{71.65} & \textbf{86.00} & \textbf{29.43} \\
        \midrule
        \multirow{2}{*}{FT} & LGIE & ~ & 0.166 & 60.11 & 0.357 & ~ & 0.124 & 71.04 & 85.47 & 29.37 \\
        ~ & MGIE & ~ & \textbf{0.163} & \textbf{61.38} & \textbf{0.348} & ~ & \textbf{0.101} & \textbf{74.79} & \textbf{87.12} & \textbf{29.68} \\
        \midrule
        \multirow{2}{*}{E2E} & LGIE & ~ & 0.144 & 64.60 & 0.327 & ~ & 0.084 & 80.90 & 88.87 & 30.10 \\
        ~ & MGIE & ~ & \textbf{0.133} & \textbf{66.25} & \textbf{0.298} & ~ & \textbf{0.082} & \textbf{82.22} & \textbf{91.14} & \textbf{30.40} \\
        \bottomrule
    \end{tabular}
    \vspace{-1.5ex}
    \captionof{table}{\textbf{Ablation study}. We attempt FZ, FT, or E2E to utilize expressive instructions. \textbf{FZ} directly treats expressive instructions as the inputs to frozen InsPix2Pix. \textbf{FT} further fine-tunes InsPix2Pix and makes it adapt to expressive instructions. Our \textbf{E2E} learns expressive instructions along with the MLLM and trains the diffusion model in an end-to-end manner.}
    \label{table:ablation}
    \vspace{-3ex}
\end{minipage}
\end{figure}

\paragraph{Baselines.}
We treat InsPix2Pix~\citep{brooks2023ins-pix2pix}, built upon the CLIP text encoder with a diffusion model for instruction-based image editing, as our baseline. We consider a similar LLM-guided image editing (LGIE) model, where LLaMA-7B~\citep{touvron2023llama} is adopted for expressive instructions $\mathcal{E}$ from instruction-only inputs but without visual perception.

\begin{figure}[h]
\centering
    \vspace{-1ex}
    \includegraphics[width=.786\linewidth]{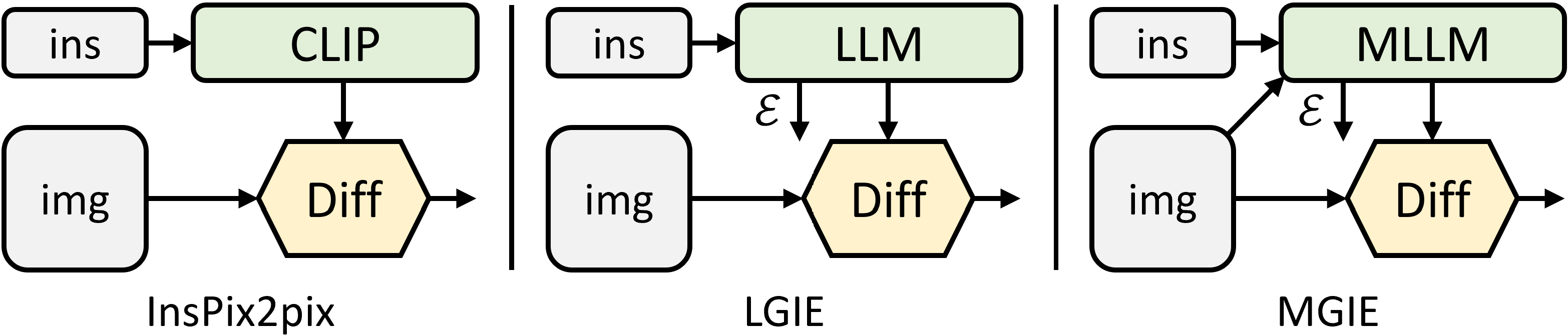}
    \vspace{-2ex}
\end{figure}

\paragraph{Implementation Details.}
The MLLM and diffusion model $\mathcal{F}$ are initialized from LLaVA-7B~\citep{liu2023llava} and StableDiffusion-v1.5~\citep{rombach2022sd}. We jointly update both for the image editing task. Note that only word embeddings and LM head in the MLLM are trainable. Following GILL~\citep{koh2023gill}, we use $N$=8 visual tokens. The edit head $\mathcal{T}$ is a 4-layer Transformer, which transforms language features into editing guidance. We adopt AdamW~\citep{loshchilov2019adamw} with the batch size of 128 to optimize MGIE. The learning rates of the MLLM and $\mathcal{F}$ are 5e-4 and 1e-4, respectively. All experiments are conducted in PyTorch~\citep{paszke2017pytorch} on 8 A100 GPUs.

\subsection{Quantitative Results}
Table~\ref{table:zero-shot} shows the zero-shot editing results, where models are trained only on IPr2Pr. For EVR and GIER that involve Photoshop-style modifications, expressive instructions can reveal concrete goals instead of brief but ambiguous commands, which makes the editing results more similar to intentions (\textit{e.g.}, higher 82.0 CVS on EVR by LGIE and higher 59.2 SSIM on GIER by MGIE). For global photo optimization on MA5k, InsPix2Pix is hard to deal with due to the scarcity of related training triples. Though trained from the same source, LGIE and MGIE can offer detailed explanations via learning with the LLM, but LGIE is still confined to its single modality. With access to images, MGIE derives explicit instructions such as \textit{which regions should brighten} or \textit{what objects are more distinct}. It can bring a significant performance boost (\textit{e.g.}, higher 66.3 SSIM and lower 0.3 photo distance). Similar results are found on MagicBrush. MGIE also achieves the best performance from the precise visual imagination and modifies the designate targets as the goals (\textit{e.g.}, higher 82.2 DINO visual similarity and higher 30.4 CTS global caption alignment).

To investigate instruction-based image editing for the specific purpose, Table~\ref{table:fine-tuned} fine-tunes models on each dataset. For EVR and GIER, all models obtain improvements after the adaptation to Photoshop-style editing tasks. Since fine-tuning makes expressive instructions more domain-specific as well, our MGIE increases the most via learning with domain-related guidance. This also helps our diffusion model to demonstrate concrete edited scenes from the fine-tuned MLLM, which benefits both global optimization and local modification (\textit{e.g.}, notably lower 0.24 LPIPS on MA5k and higher 95.3 CVS on MagicBrush). MGIE is consistently superior to LGIE in all aspects of editing since our visual-aware guidance is more aligned with the intended goal. From the above experiments, we illustrate that learning with expressive instructions can effectively enhance image editing, and visual perception plays a crucial role in deriving explicit guidance for the greatest enhancements.

\paragraph{Trade-off between $\alpha_\mathcal{X}$ and $\alpha_\mathcal{V}$.}
There are two goals in image editing: manipulate the target as the instruction and preserve the remaining as the input image. Fig.~\ref{fig:trade-off} plots the trade-off curves between the instruction ($\alpha_\mathcal{X}$) and input consistency ($\alpha_\mathcal{V}$). We fix $\alpha_\mathcal{X}$ as 7.5 and vary $\alpha_\mathcal{V}$ in $[1.0, 2.2]$. Higher $\alpha_\mathcal{V}$ will make an editing result more similar to the input but less aligned with the instruction. X-axis calculates the CLIP directional similarity as how much the editing follows the instruction; Y-axis is the feature similarity to the input image from the CLIP visual encoder. Through concrete expressive instructions, we surpass InsPix2Pix in all settings. Our MGIE additionally results in comprehensive enhancements by learning with explicit visual-related guidance. This supports robust improvement, whether requiring higher input correlation or edit relevance. 

\begin{figure}[t]
\begin{minipage}{.3\textwidth} \vspace{-\topskip}
\centering
    \includegraphics[width=\linewidth]{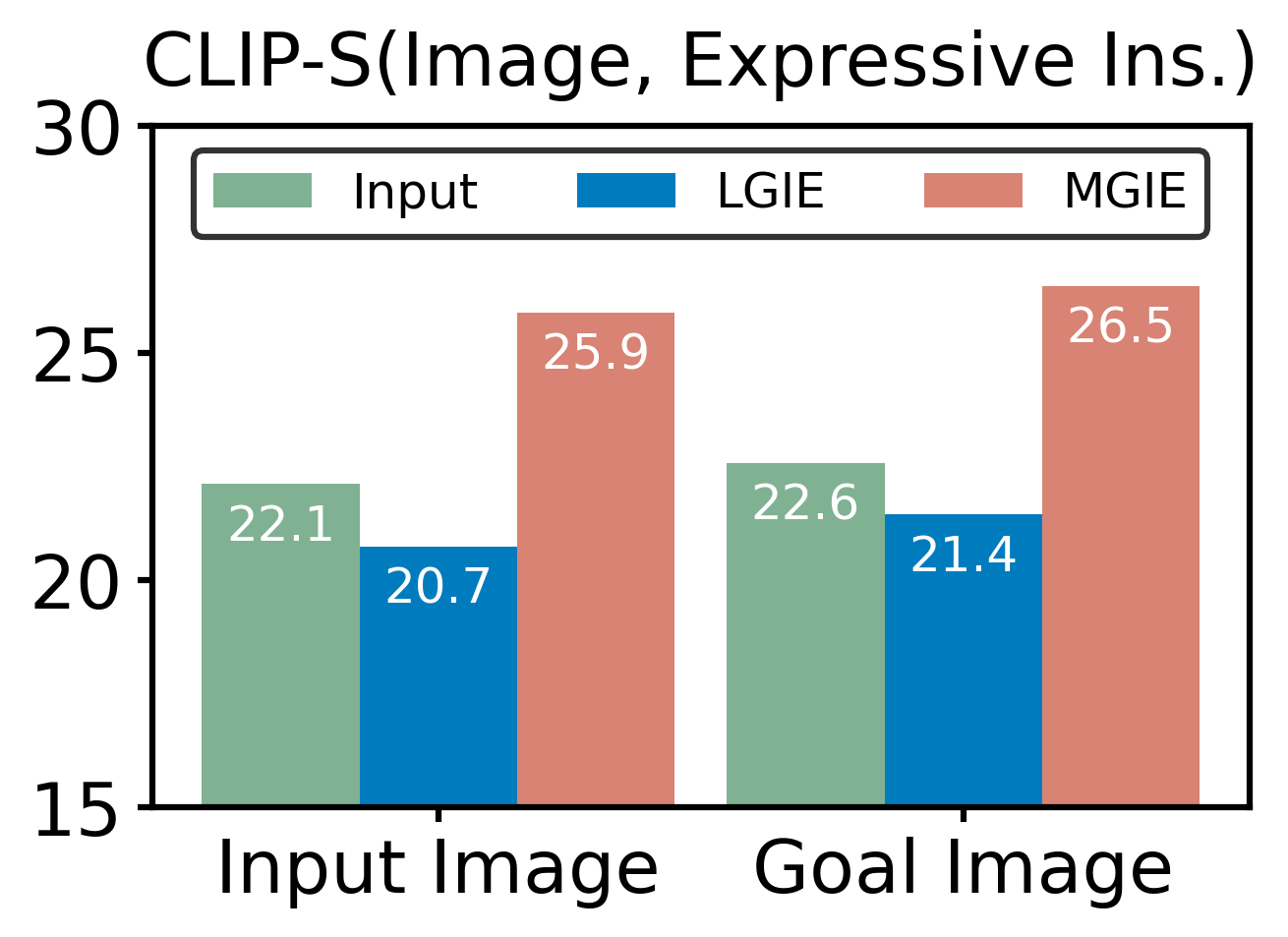}
    \vspace{-3.5ex}
    \captionof{figure}{\textbf{CLIP-S} across images (input / goal) and expressive instructions.}
    \label{fig:mllm-help}
    \vspace{-3ex}
\end{minipage}~~
\begin{minipage}{.3\textwidth} \vspace{-\topskip}
\centering
    \vspace{-0.6ex}
    \includegraphics[width=\linewidth]{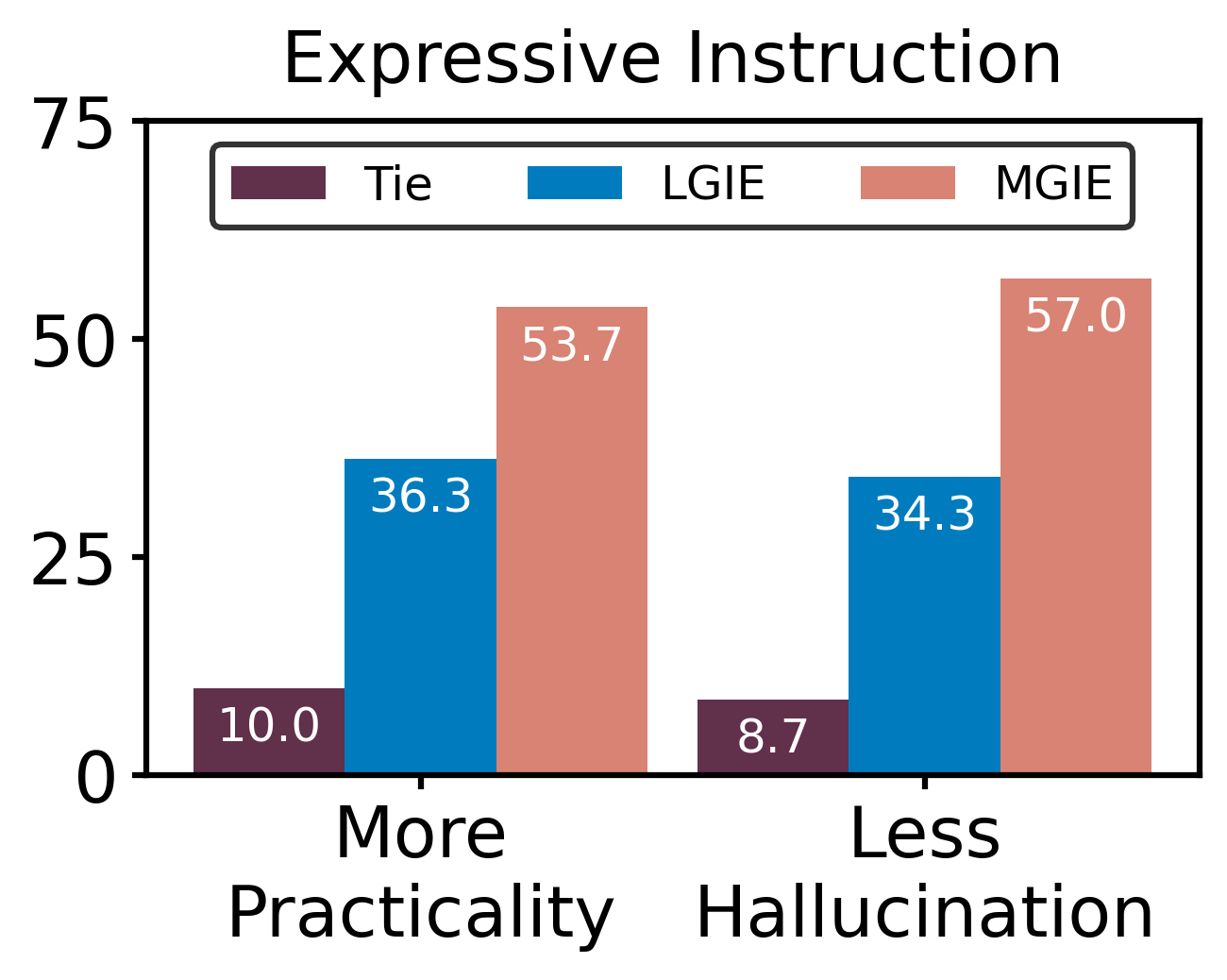}
    \vspace{-3.5ex}
    \captionof{figure}{\textbf{Human eval} of expressive instructions quality.}
    \label{fig:he-ins}
    \vspace{-3ex}
\end{minipage}~~
\begin{minipage}{.37\textwidth}
\centering
    \includegraphics[width=\linewidth]{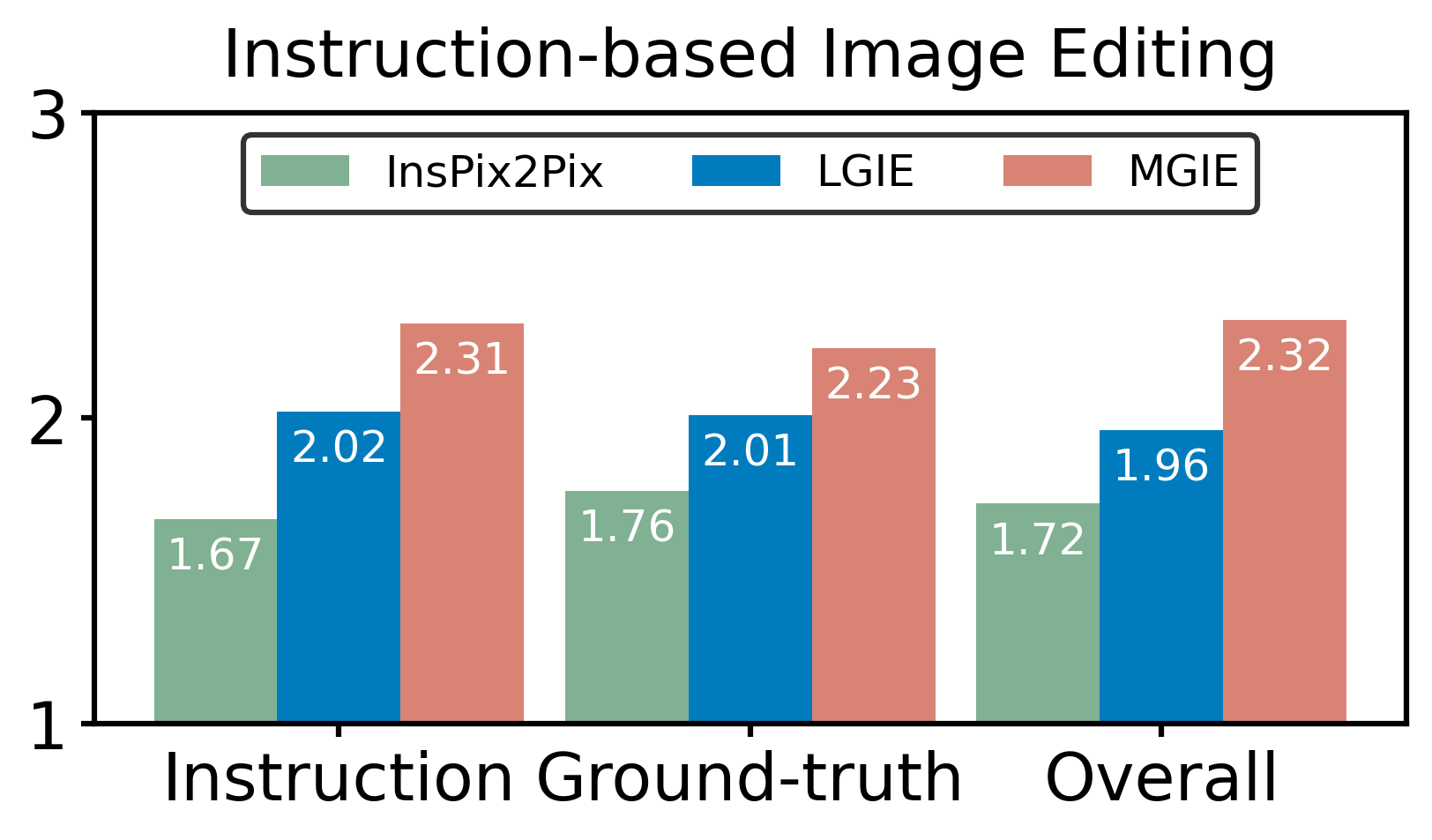}
    \vspace{-3.5ex}
    \captionof{figure}{\textbf{Human eval} of image editing results in terms of instruction following, ground-truth relevance, and overall quality.}
    \label{fig:he-ie}
    \vspace{-3ex}
\end{minipage}
\end{figure}

\subsection{Ablation Study} \label{sec:ablation-study}
MLLM-Guided Image Editing exhibits encouraging improvement in both zero-shot and fine-tuning scenarios. Now, we investigate different architectures to use expressive instructions. Table~\ref{table:ablation} considers \textbf{FZ}, \textbf{FT}, and our \textbf{E2E}. FZ directly uses the derived expressive instructions\footnote{During the ablation study, we employ concise summarized expressive instructions for a fair comparison.} as the input prompts to the frozen InsPix2Pix. In spite of having additional guidance, the scenario still differs from the trained editing instructions, which makes it difficult to deal with. LGIE even hurts the performance as it may mislead due to the shortage of visual perception. FT fine-tunes InsPixPix and adapts it to expressive instructions. These results support that image editing can benefit from explicit guidance along the derivation of instructions from the LLM/MLLM. E2E updates the editing diffusion model in conjunction with the LM, which learns to extract applicable guidance and discard irrelevant narration simultaneously through the end-to-end hidden states. In addition, our E2E can also avoid the potential error that may be propagated from the expressive instructions. Hence, we observe the most enhancements in both global optimization (MA5k) and local editing (MagicBrush). Among FZ, FT, and E2E, MGIE consistently surpasses LGIE. This indicates that expressive instructions with crucial visual perception are always advantageous across all ablation settings.

\begin{figure}[t]
\centering
    \includegraphics[width=\linewidth]{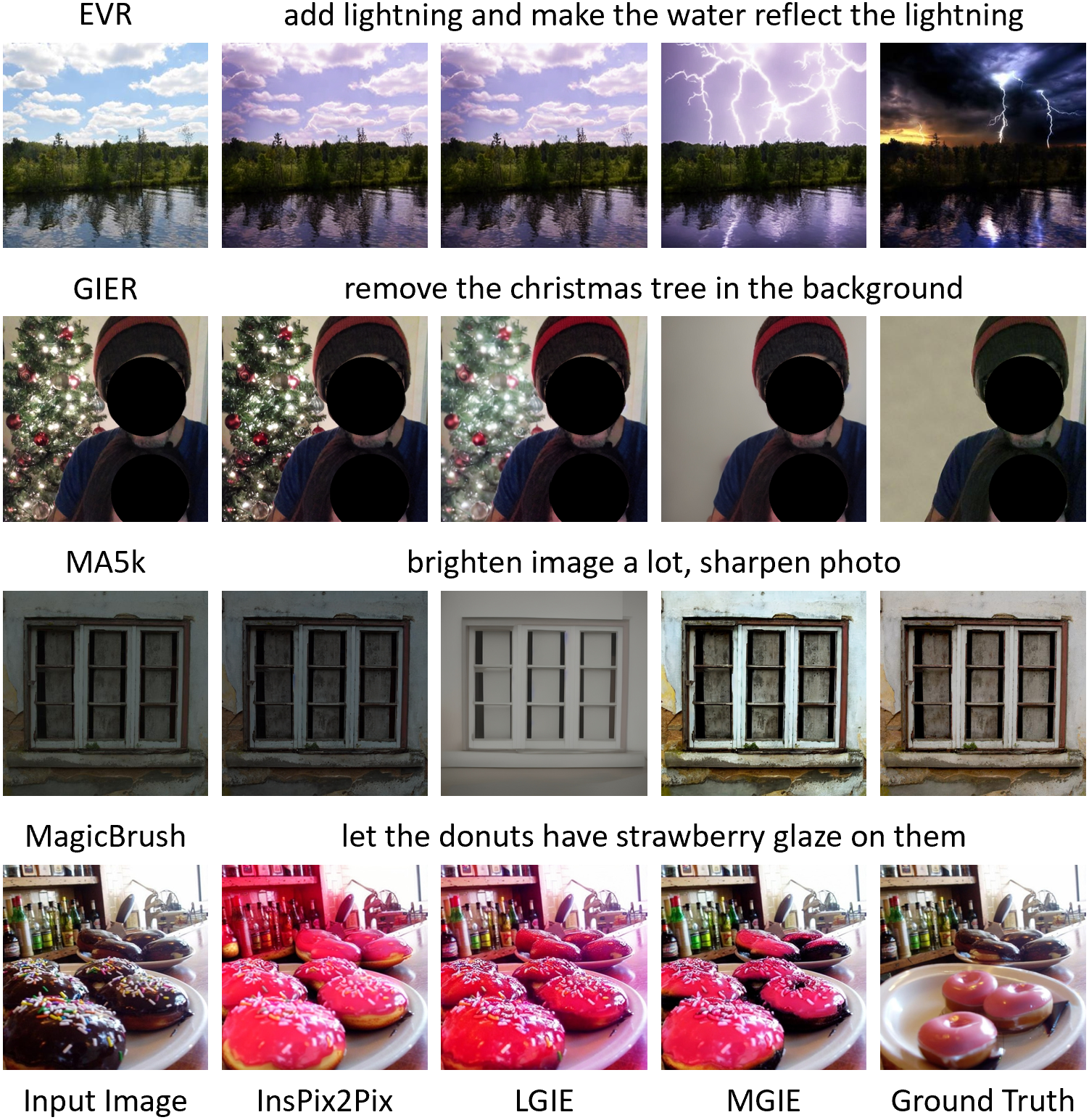}
    \vspace{-3ex}
    \caption{\textbf{Qualitative comparison} between InsPix2Pix, LGIE, and our MGIE. For the 1st example, MGIE can showcase the clear ``\textit{lightning}'' in the sky and its reflection on the water. For the 2nd one, although LGIE accurately targets the Christmas tree, only MGIE removes it in the background. For photo optimization (the 3rd example), InsPix2Pix fails to adjust the brightness, and LGIE makes the whole photo white and obviously distinct. In contrast, MGIE follows the instruction to brighten as well as sharpen it. Moreover, in the 4th one, MGIE puts the ``\textit{glaze}'' only on the donuts, but baselines even draw the entire image in strawberry pink.}
    \label{fig:qual_comparison}
    \vspace{-3ex}
\end{figure}

\paragraph{Why MLLM Guidance is Helpful?}
Fig.~\ref{fig:mllm-help} presents the CLIP-Score between input or ground-truth goal images and expressive instructions. A higher CLIP-S to input images indicates that instructions are relevant to the editing source. Better alignment with goal images provides explicit and correlated edit guidance. Without access to visual perception, expressive instructions from LGIE are limited to general language imagination, which is not tailored to the source image. The CLIP-S are even lower than the original instructions. By contrast, MGIE is more aligned with inputs/goals, which explains why our expressive instructions are helpful. With a clear narration of the intended result, our MGIE can achieve the greatest improvements in image editing.

\paragraph{Human Evaluation.}
Apart from automatic metrics, we conduct a human evaluation to study generated expressive instructions and image editing results. We randomly sample 25 examples for each dataset (100 in total) and consider humans to rank across baselines and MGIE. To avoid potential ranking bias, we hire 3 annotators for each example. Fig.~\ref{fig:he-ins} plots the quality of generated expressive instructions. Precise guidance is informative and aligns with the intended goal (More Practicality). At the same time, it should avoid incorrect or unrelated explanations (Less Hallucination). Firstly, over 53\% support that MGIE provides more practical expressive instructions, which facilitates the image editing task with explicit guidance. Meanwhile, 57\% of labelers indicate that our MGIE can prevent irrelevant descriptions from language-derived hallucinations in LGIE since it perceives the image to have a precise goal for editing. Fig.~\ref{fig:he-ie} compares the image editing results by InsPix2Pix, LGIE, and our MGIE in terms of instruction following, ground-truth relevance, and overall quality. The ranking score is ranging from 1 to 3, higher is better. With derived expressive instructions from the LLM or MLLM, LGIE and MGIE both outperform the baseline and perform image editing that is correlated with the instruction as well as similar to the ground-truth goal. Additionally, since our expressive instructions can provide concrete and visual-aware guidance, MGIE has the best human preference in all aspects, including the overall editing quality. These performance trends also align with automatic evaluations, which support our usage of metrics.

\begin{wrapfigure}{r}{0.2\textwidth}
\centering \tablestyle{1pt}{1.1}
    \vspace{-.8\intextsep}
    \begin{tabular}{ccc}
        \toprule
        BS & InsPix2Pix & MGIE \\
        \midrule
        1 & 6.8 & 9.2 \\
        4 & 16.5 & 20.6 \\
        8 & 31.5 & 36.9 \\
        \bottomrule
    \end{tabular}
    \vspace{-2ex}
    \captionof{table}{Time cost.}
    \label{table:time}
    \vspace{-\intextsep}
\end{wrapfigure}

\paragraph{Inference Efficiency.}
Despite relying on MLLM to facilitate image editing, MGIE only rolls out concise expressive instructions (less than 32 tokens) and contains feasible efficiency as InsPix2Pix. Table~\ref{table:time} presents the inference time cost on an NVIDIA A100 GPU. For a single input, MGIE can accomplish the editing task in 10 seconds. With greater data parallelization, we take a similar amount of time (\textit{e.g.}, 37 seconds when batch size 8). The entire process can be affordable in one GPU (40GB). In summary, our MGIE surpasses the baseline on quality yet maintains competitive efficiency, leading to effective and practical image editing.

\begin{figure}[t]
\centering
    \includegraphics[width=\linewidth]{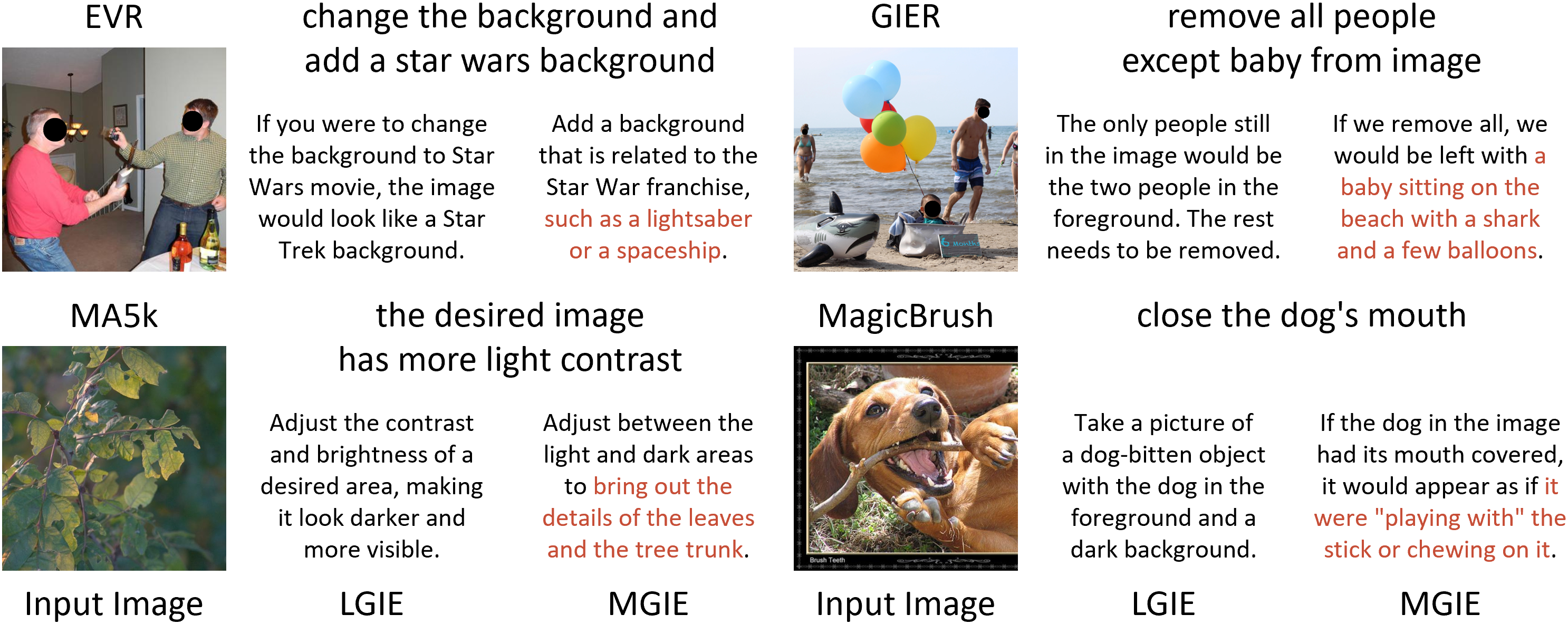}
    \vspace{-3ex}
    \caption{\textbf{Qualitative comparison} of expressive instructions by LGIE and our MGIE. Due to the limitation of the single modality, LGIE can only have language-based insight but may derive irrelevant or even wrong explanations for image editing (\textit{e.g.}, ``\textit{two people still in the foreground}'' for GIER). With access to images, MGIE provides explicit visual imagination after the editing such as ``\textit{baby on the beach with a shark}'' or ``\textit{bring out details of leaves and trunk}''. More surprisingly, we can link ``\textit{lightsaber or spaceship}'' from Star Wars and describe ``\textit{chewing on the stick}'' for the dog, which is aligned with the intended goal.}
    \label{fig:qual_ins}
    \vspace{-3ex}
\end{figure}

\paragraph{Qualitative Comparisons.}
Fig.~\ref{fig:qual_comparison} illustrates the visualized comparison on all used datasets. Fig.~\ref{fig:qual_ins} further compares the expressive instructions by LGIE or MGIE. Our superior performance benefits from the explicit guidance of visual-related expressive instructions. Please visit our project website\footnote{Project website:~\url{https://mllm-ie.github.io}\label{foot:project}} for more qualitative results.

\section{Conclusion}
We propose MLLM-Guided Image Editing (MGIE) to enhance instruction-based image editing via learning to produce expressive instructions. Instead of brief but ambiguous guidance, MGIE derives explicit visual-aware intention and leads to reasonable image editing. We conduct extensive studies from various editing aspects and demonstrate that our MGIE effectively improves performance while maintaining competitive efficiency. We also believe the MLLM-guided framework can contribute to future vision-and-language research.

\clearpage

\bibliography{iclr2024_conference}
\bibliographystyle{iclr2024_conference}

\clearpage

\appendix

\begin{figure}[t]
\begin{minipage}{.48\textwidth}
\centering \tablestyle{1.2pt}{1.1}
    \begin{tabular}{cccccccc}
        \toprule
        \textbf{Method} & ~ & \multicolumn{2}{c}{\textbf{MA5k}} & ~ & \multicolumn{3}{c}{\textbf{MagicBrush}} \\
        \cmidrule{3-4} \cmidrule{6-8}  ~ & ~ & SSIM$\uparrow$ & LPIPS$\downarrow$ & ~ & DINO$\uparrow$ & CVS$\uparrow$ & CTS$\uparrow$ \\
        \midrule
        InsPix2Pix & ~ & 58.92 & 0.359 & ~ & 71.46 & 85.22 & \underline{29.34} \\
        + Enc$_\text{LLaMA}$ & ~ & \underline{59.08} & \textbf{0.334} & ~ & \underline{72.38} & \underline{85.99} & 29.29 \\
        + Enc$_\text{LLaVA}$ & ~ &  \textbf{60.94} & \underline{0.352} & ~ & \textbf{74.10} & \textbf{87.21} & \textbf{29.37} \\
        \midrule
        HIVE & ~ & \underline{65.17} & \underline{0.302} & ~ & 78.95 & 88.23 & 29.42 \\
        InsEdit & ~ & 59.59 & 0.364 & ~ & \textbf{83.26} & \textbf{91.16} & \underline{29.80} \\
        MGIE & ~ & \textbf{66.25} & \textbf{0.298} & ~ & \underline{82.22} & \underline{91.14} & \textbf{30.40} \\
        \bottomrule
    \end{tabular}
    \vspace{-1.5ex}
    \captionof{table}{\textbf{Zero-shot editing comparison} to different instruction encoders (Enc), human feedback (HIVE), and mask-then-inpaint (InsEdit).}
    \label{table:more-baseline}
    \vspace{-3ex}
\end{minipage}~~
\begin{minipage}{.50\textwidth}
\centering \tablestyle{1.2pt}{1.1}
    \begin{tabular}{ccccccccc}
        \toprule
        \textbf{Method} & \textbf{Size} & ~ & \multicolumn{2}{c}{\textbf{MA5k}} & ~ & \multicolumn{3}{c}{\textbf{MagicBrush}} \\
        \cmidrule{4-5} \cmidrule{7-9} ~ & ~ & ~ & SSIM$\uparrow$ & LPIPS$\downarrow$ & ~ & DINO$\uparrow$ & CVS$\uparrow$ & CTS$\uparrow$ \\
        \midrule
        \multicolumn{2}{c}{InsPix2Pix} & ~ & 58.92 & 0.359 & ~ & 71.46 & 85.22 & 29.34 \\
        \midrule
        \multirow{2}{*}{LGIE} & 7B & ~ & \textbf{64.60} & 0.327 & ~ & \textbf{80.90} & \textbf{88.87} & 30.10 \\
        ~ & 13B & ~ & 63.50 & \textbf{0.308} & ~ & 80.18 & 88.77 & \textbf{30.31} \\
        \midrule
        \multirow{3}{*}{MGIE} & 6.7B & ~ & 63.78 & 0.300 & ~ & 78.82 & 90.01 & 29.47 \\
        ~ & 7B & ~ & \textbf{66.25} & \underline{0.298} & ~ & \textbf{82.22} & \underline{91.14} & \underline{30.40} \\
        ~ & 13B & ~ & \underline{65.91} & \textbf{0.279} & ~ & \underline{82.15} & \textbf{91.52} & \textbf{30.75} \\
        \bottomrule
    \end{tabular}
    \vspace{-1.5ex}
    \captionof{table}{\textbf{Zero-shot editing comparison} of different LM sizes. We treat the visual-tuned OPT-6.7B in our used MGIE-6.7B.}
    \label{table:size}
    \vspace{-3ex}
\end{minipage}
\end{figure}

\section{Additional Results}
\paragraph{Comparison to More Baselines.}
InsPix2Pix~\citep{brooks2023ins-pix2pix} applies the CLIP encoder~\citep{radford2021clip}, which is insufficient to capture the transformation for editing. We treat the stronger LLM/MLLM as the instruction encoder (Enc) and follow the same training strategy. Table~\ref{table:more-baseline} presents that adopting LLaMA~\citep{touvron2023llama} / LLaVA~\citep{liu2023llava} can slightly outperform CLIP, and the visual-aware encoding is also crucial in the original InsPix2Pix. However, they still contain a performance gap with our MGIE, which indicates that merely replacing the instruction encoder is not enough for their limitation. We further consider HIVE~\citep{zhang2023hive} and InsEdit~\citep{wang2023ins-edit} for the additional baselines. HIVE collects human preference and enhances InsPix2Pix via reward feedback learning. InsEdit depends on an external segmentation model to provide the target mask and performs inpainting as the editing result. The results demonstrate that MGIE consistently surpasses HIVE without extra human feedback, which is more data-efficient for training. InsEdit is superior in local editing with its mask-then-inpaint but not in global optimization. The mask should always be the entire photo, and the inpainting is not capable of adjusting the brightness or saturation. In contrast, through learning with the derivation of the MLLM, our MGIE performs robustly in both.

\begin{figure}[h]
\centering
    \vspace{-2ex}
    \includegraphics[width=\linewidth]{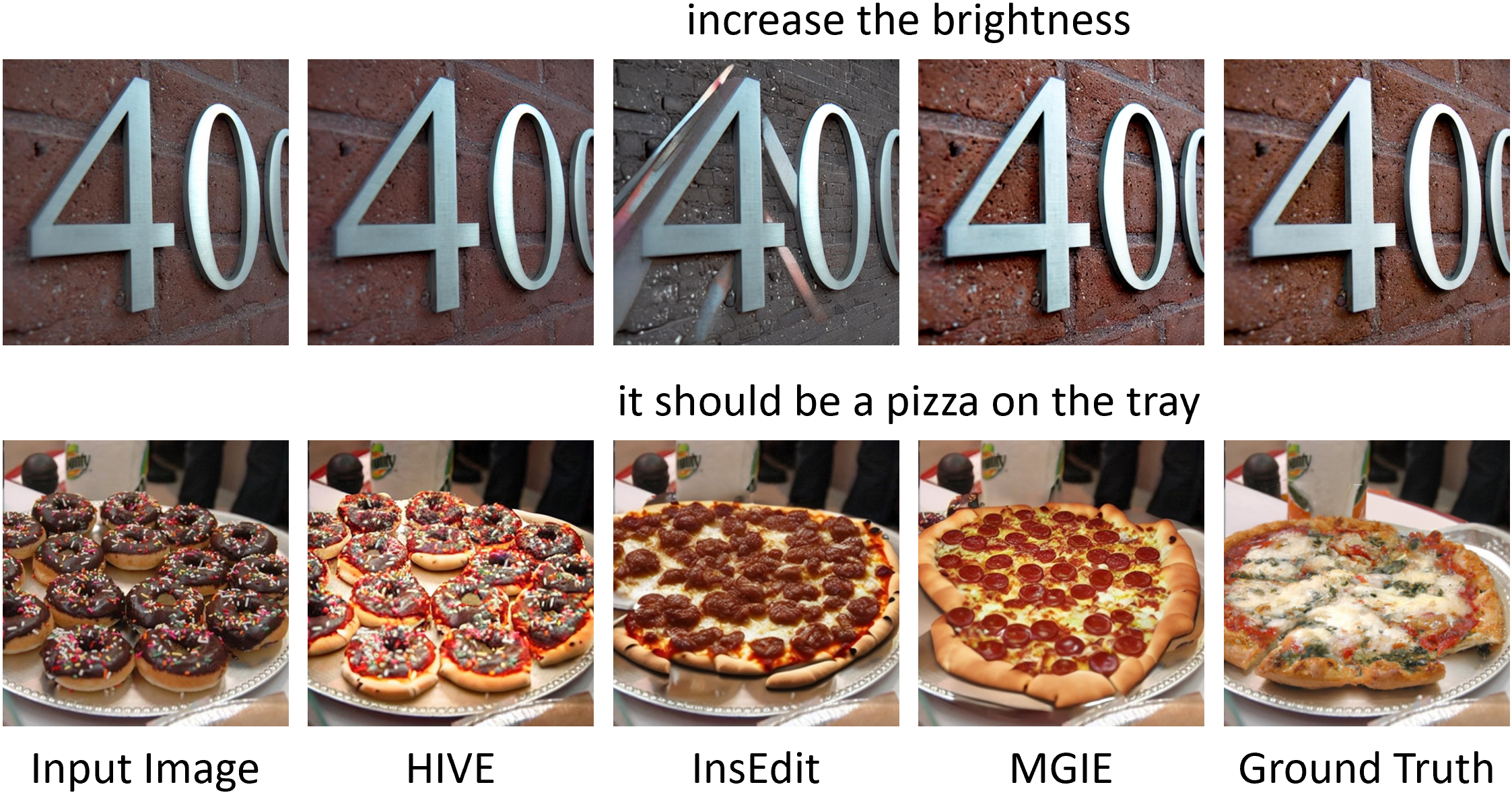}
    \vspace{-6ex}
\end{figure}

\begin{figure}[t]
\begin{minipage}{.32\textwidth}
\centering
    \includegraphics[width=\linewidth]{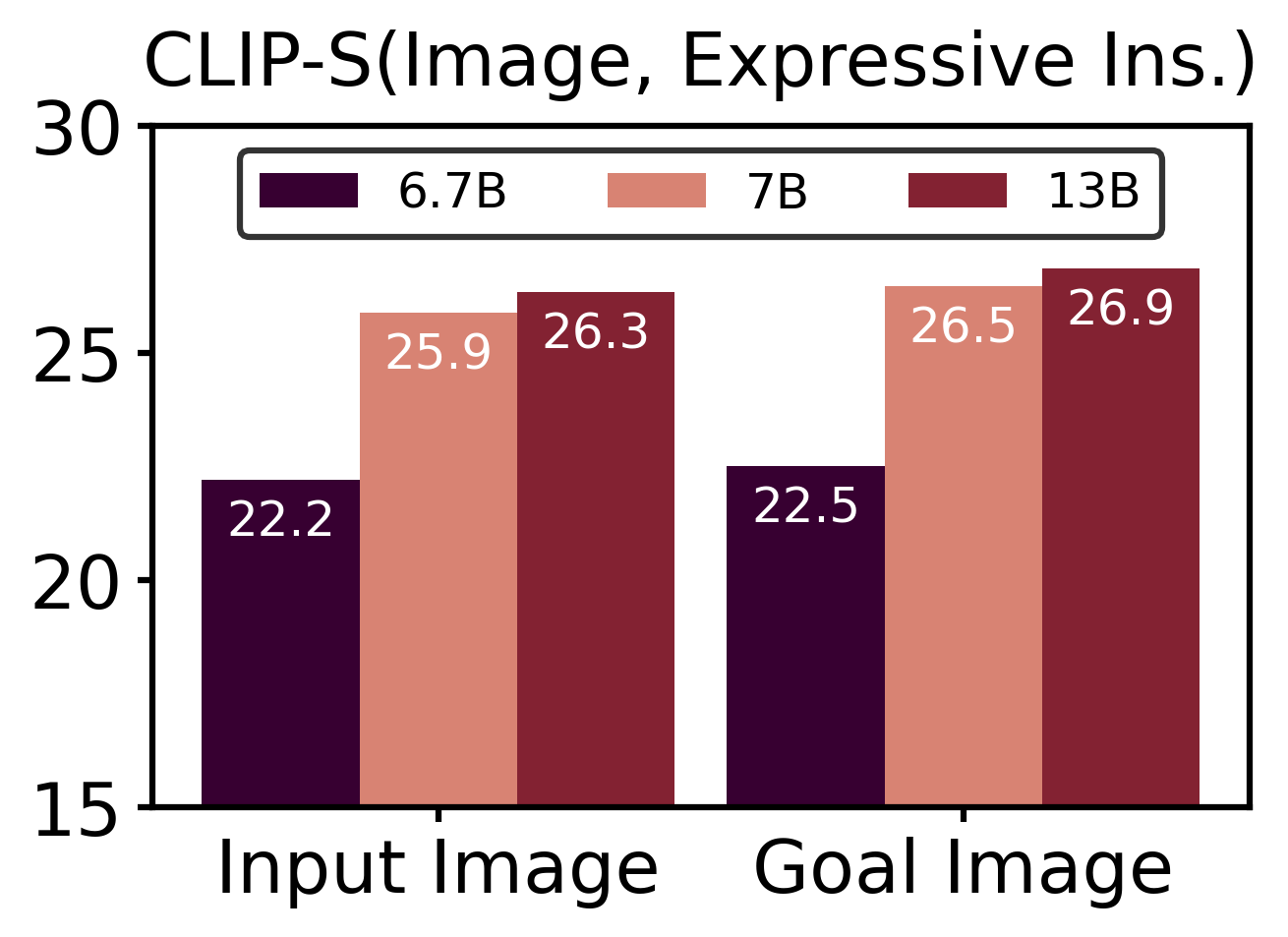}
    \vspace{-3.5ex}
    \captionof{figure}{\textbf{CLIP-S} across images and expressive instructions by different sizes of MGIE.}
    \label{fig:size}
    \vspace{-3ex}
\end{minipage}~~
\begin{minipage}{.32\textwidth}
\centering
    \includegraphics[width=\linewidth]{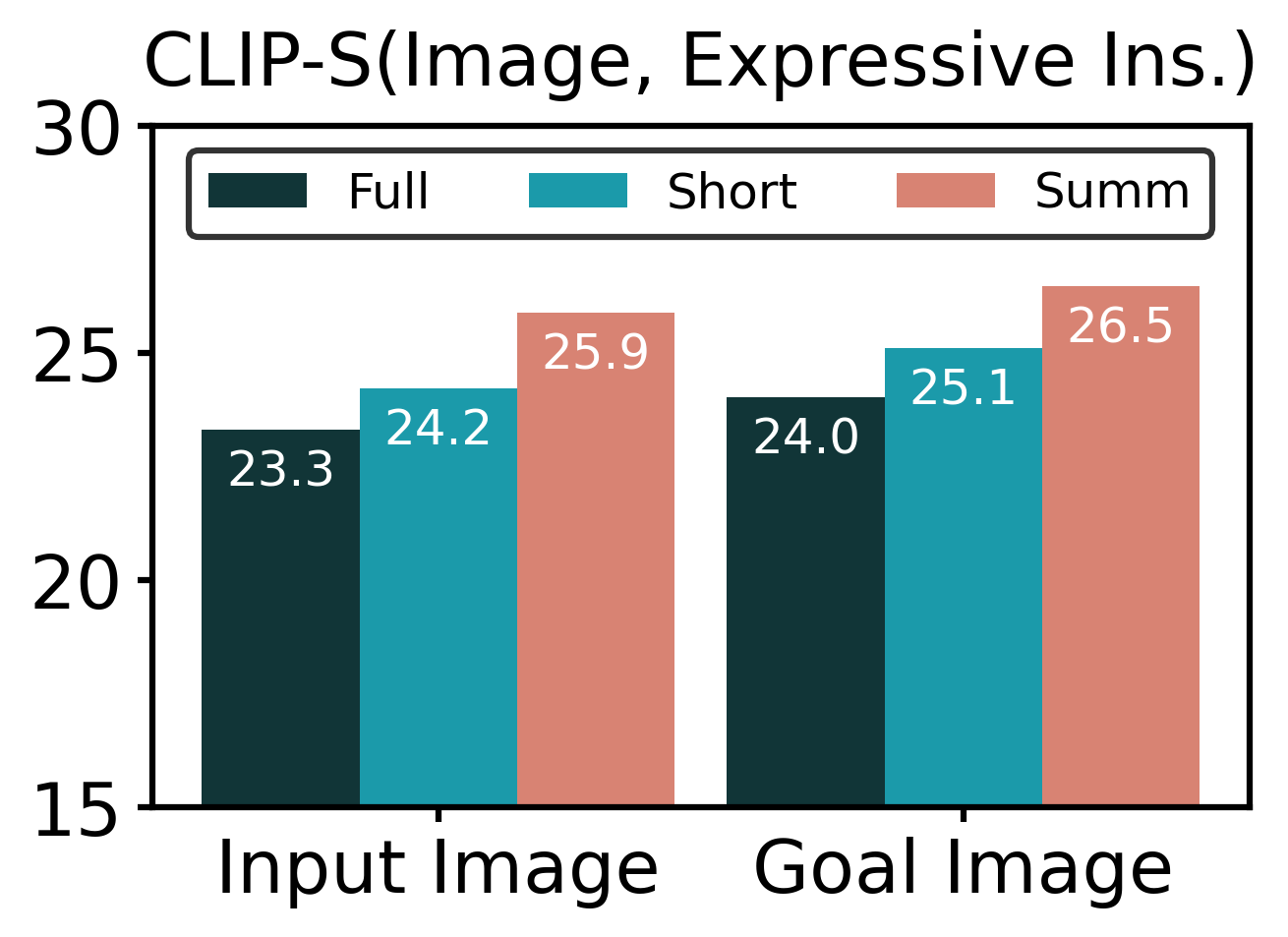}
    \vspace{-3.5ex}
    \captionof{figure}{\textbf{CLIP-S} across images and expressive instructions (full / short / summarized).}
    \label{fig:full-short-summ}
    \vspace{-3ex}
\end{minipage}~~
\begin{minipage}{.32\textwidth}
\centering
    \includegraphics[width=\linewidth]{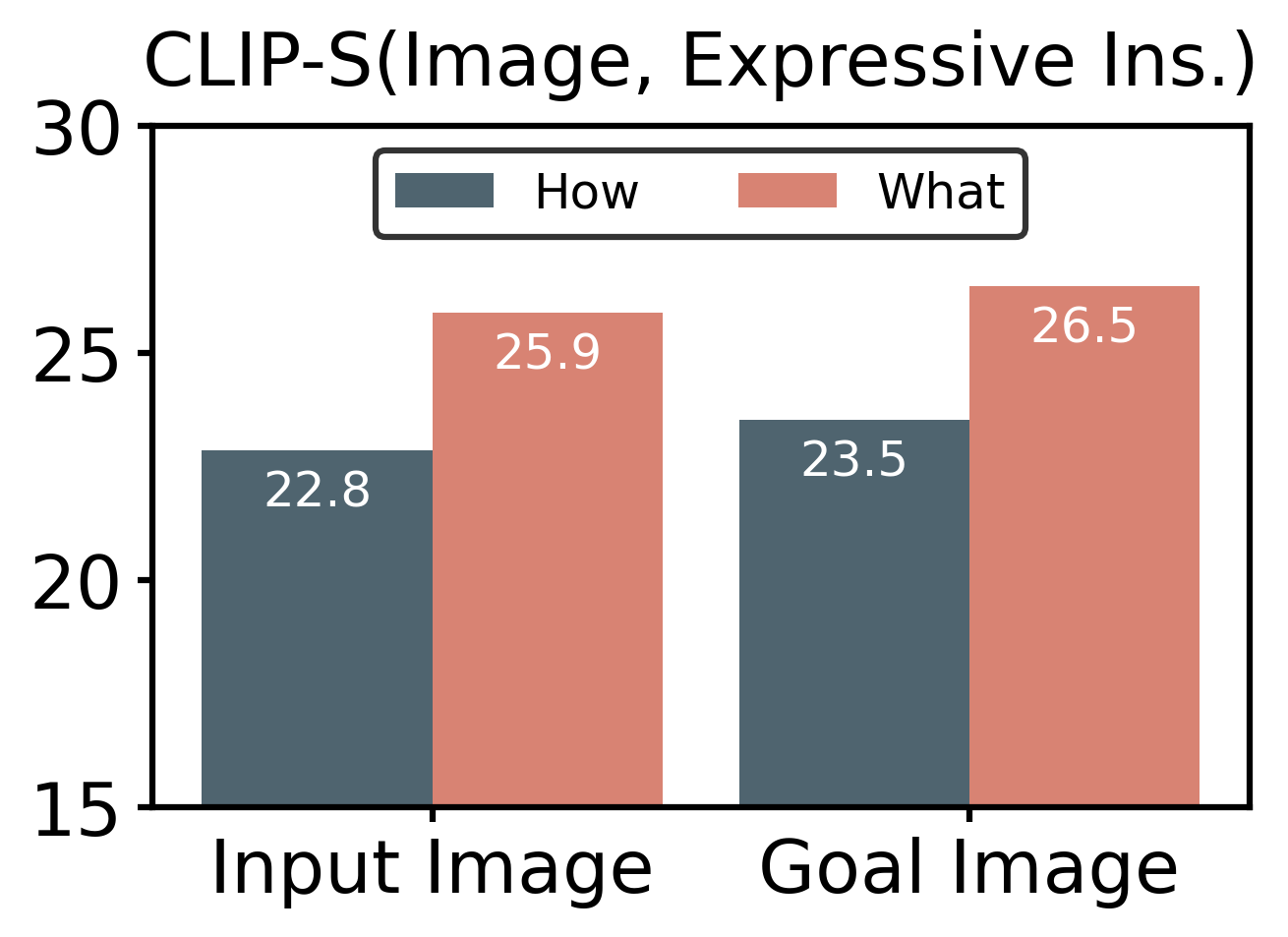}
    \vspace{-3.5ex}
    \captionof{figure}{\textbf{CLIP-S} across images and expressive instructions by the ``\textit{how}'' or ``\textit{what}'' prompt.}
    \label{fig:how-what}
    \vspace{-3ex}
\end{minipage}
\end{figure}

\paragraph{Does Larger LM Help?}
Our MGIE leverages LLMs/MLLMs to enhance instruction-based image editing. We investigate that if stronger LMs can bring more improvement. We consider the visual-tuned OPT-6.7B~\citep{zhang2022opt} and the larger LLaVA-13B in Table~\ref{table:size}. We also adopt LLaMA-13B for LGIE. Even though MGIE-7B has a similar size to MGIE-6.7B, its LLaVA is more powerful than OPT, which leads to an accurate visual imagination for better editing. The 13B obtains further performance gain for both LGIE and MGIE. Fig.~\ref{fig:size} plots the CLIP-Score of expressive instructions by different sizes of MGIE. This indicates that the guidance from larger LMs is more alignment with the vision, and thus can benefit image editing more.

\paragraph{Learning with Summarized Expressive Instruction.}
By default, MGIE learns with summarized expressive instructions for better performance and inference efficiency. We compare our form to the full description and the one making ``\textit{what will this image be like if }\texttt{[INS]}\textit{ (in short)}'' as the prompt. Fig.~\ref{fig:full-short-summ} illustrates that Full is not that aligned with images due to its irrelevant narrations (\textit{e.g.}, ``\textit{filled with people enjoying the waterfront}''). Although Short can derive brief statements (21.1 tokens), our\\Summ (22.7 tokens) is still more aligned with input or goal images. In the qualitative aspect, Short's ``\textit{create a dangerous element}'' is not explicit for ``\textit{add a shark}''. Short even merely captions the photo but without ``\textit{in Seattle}''. In contrast, our Summ provides concise yet concrete guidance, such as ``\textit{a shark swimming on the lake}'' or ``\textit{iconic Space Needle, urban setting}''.

\begin{figure}[h]
\centering
    \vspace{-1ex}
    \includegraphics[width=.838\linewidth]{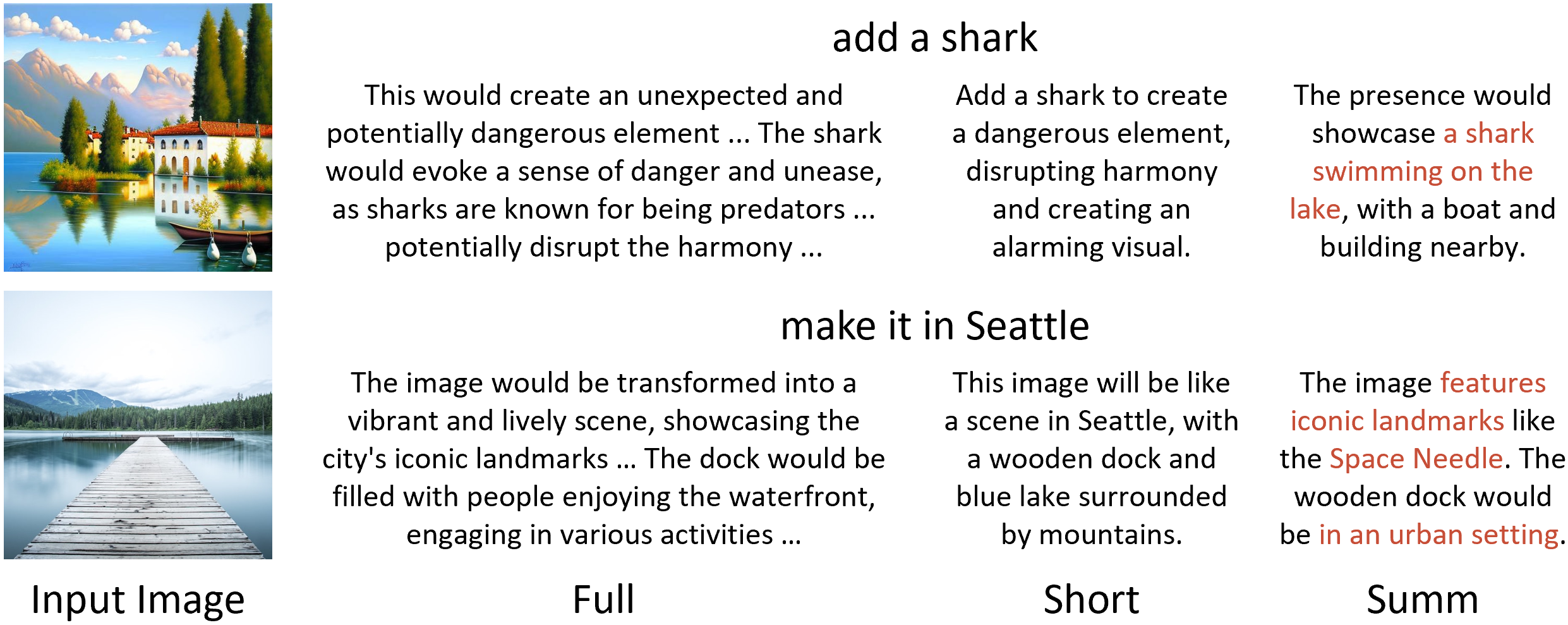}
    \vspace{-1ex}
\end{figure}

Apart from the used ``\textit{What}'' prompt, we also investigate a ``\textit{How}'' prompt as ``\textit{how to edit this image and} \texttt{[ins]}'' for expressive instructions. Fig.~\ref{fig:how-what} shows that our ``\textit{What}'' is more aligned, which can guide image editing with more relevant visual implications, such as ``\textit{painted in hues of red, orange, and yellow}'' for Autumn or ``\textit{famous landmarks as Kremlin}'' for Russia. ``\textit{How}'' miscomprehends the instruction as ``\textit{replace the whole garden with a beach}''. However, it should only manipulate the end of the stairs yet remain ``\textit{the stairway surrounded by lush greenery}''.

\begin{figure}[h]
\centering
    \vspace{-1ex}
    \includegraphics[width=\linewidth]{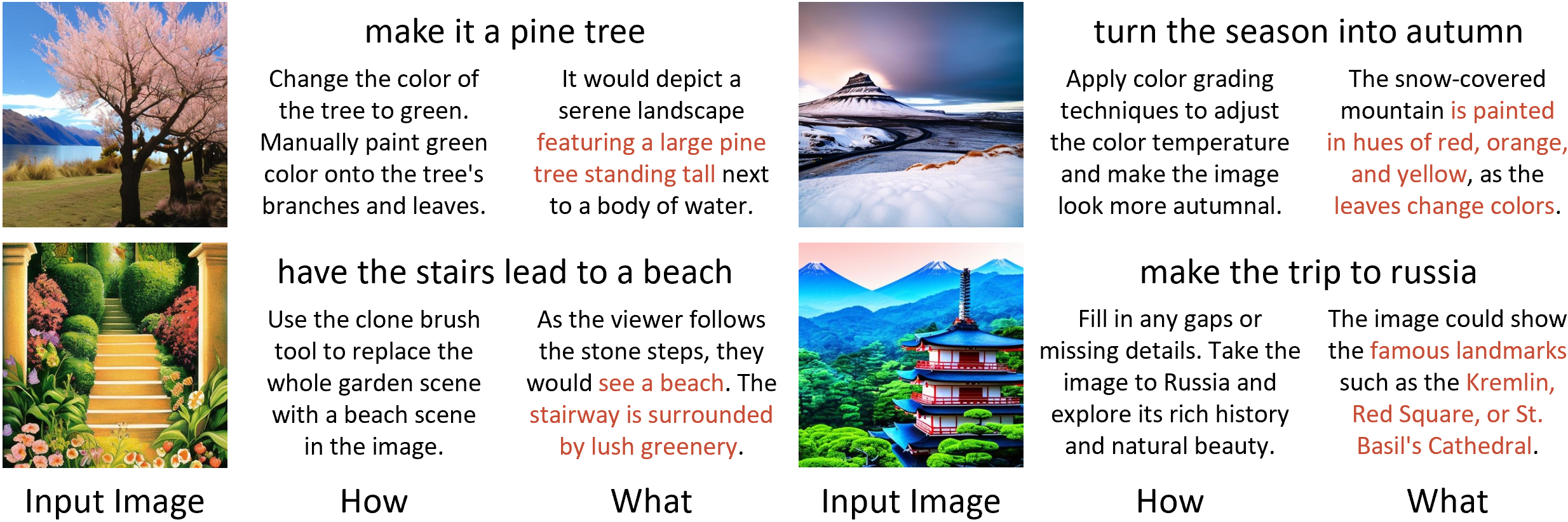}
    \vspace{-4ex}
\end{figure}

\paragraph{How Many Visual Tokens do We Need?}
Our editing head projects the guidance modality from the MLLM to the diffusion model. We follow GILL~\citep{koh2023gill} and apply $N$=8 visual tokens by default. Here we investigate the effectiveness of different numbers of \texttt{[IMG]}. The results indicate that less \texttt{[IMG]} makes the extracted visual imagination insufficient for effective guidance, resulting in a significant performance drop. While more \texttt{[IMG]} can bring further enhancements, we also find that the performance gets similar when using more than 4 \texttt{[IMG]}.

\begin{figure}[h]
\centering
    \vspace{-1ex}
    \includegraphics[width=.6748\linewidth]{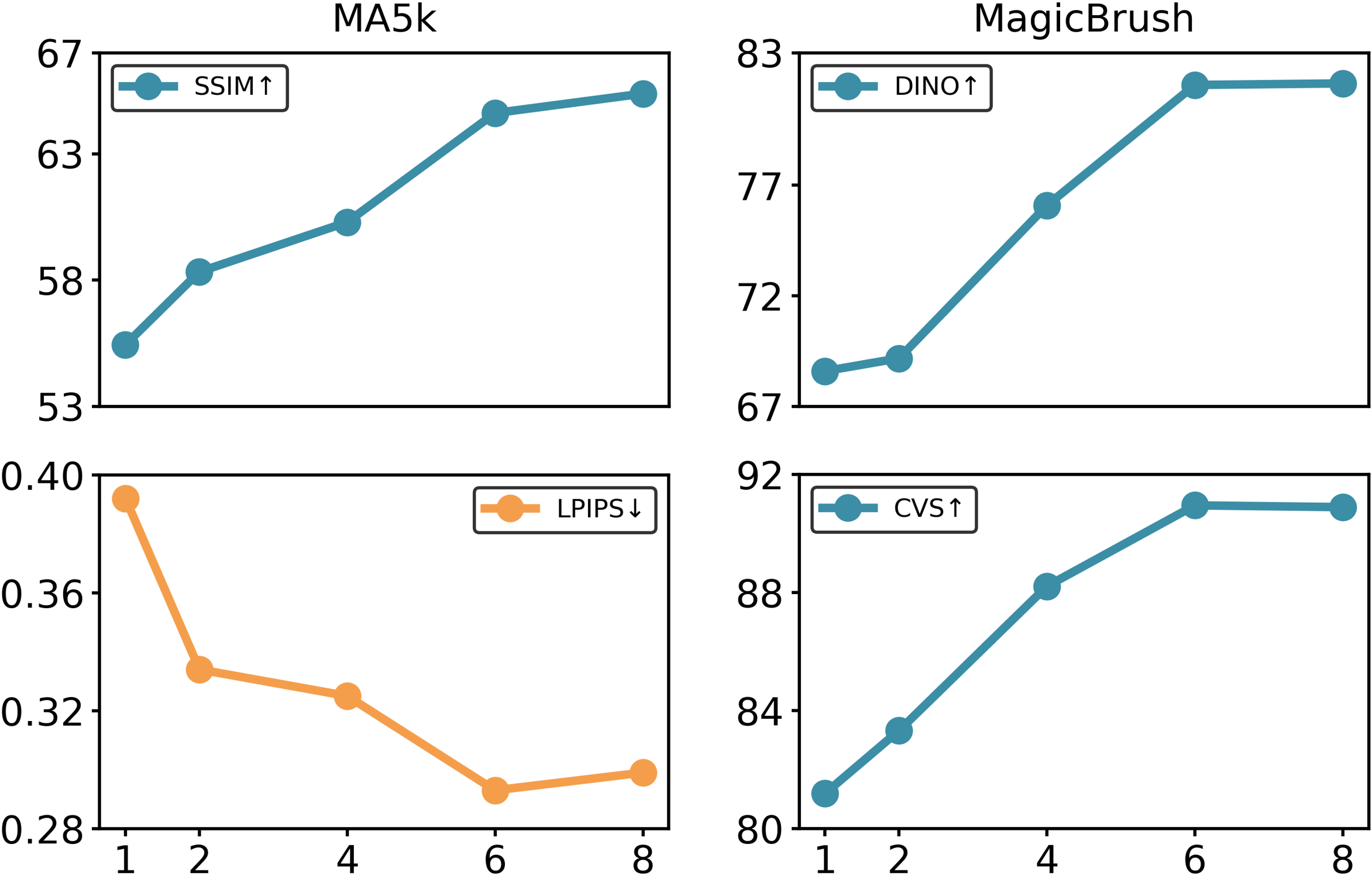}
    \vspace{-1ex}
\end{figure}

\paragraph{Qualitative Results of Different $\alpha_\mathcal{V}$.}
MGIE adopts the weight $\alpha_\mathcal{V}$ to adjust the level of editing. A higher $\alpha_\mathcal{V}$ makes the editing result more similar to the input, while a lower $\alpha_\mathcal{V}$ leads to more editing applied onto the image. Hence we can control the extent of visual transformation for both local (\textit{e.g.,} the color of cherries) and global editing (\textit{e.g.,} the style of the painting).

\begin{figure}[h]
\centering
    \vspace{-1ex}
    \includegraphics[width=\linewidth]{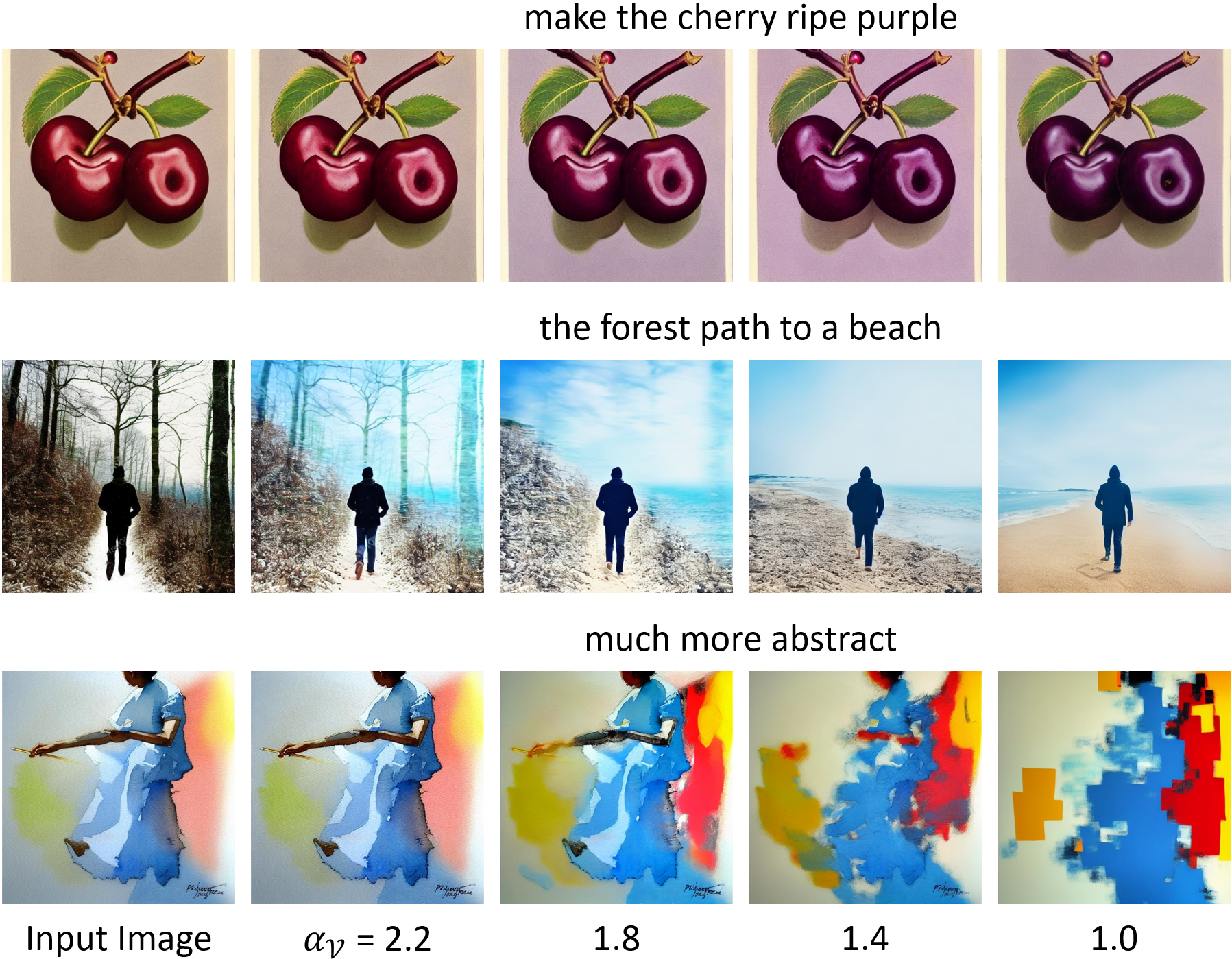}
    \vspace{-4ex}
\end{figure}

\paragraph{Comparison to Description-based Baselines.}
In addition to instruction-based baselines, we also consider description-based editing models. We leverage GIT~\citep{wang2022git} to caption the input image as its input description and ChatGPT to merge the edit instruction as the goal description via the prompt ``\textit{Combine two sentences A:}~\texttt{[description]}~\textit{and B:}~\texttt{[instruction]}~\textit{into a single sentence. The output should be at most similar to sentence A}''. For instance, ``\textit{a girl is walking at the beach}'' and ``\textit{give her a hat}'' will be transformed into ``\textit{a girl with a hat is walking at the beach}''. For MagicBrush, we directly apply their released descriptions instead. Text2LIVE~\citep{bar-tal2022text2live} and Null-Inv~\citep{mokady2023null-text} only yield feasible results on the traditional L1 distance but are obviously inferior to our MGIE on semantic-level evaluations (\textit{e.g.,} lower CVS), which supports that they cannot present concrete editing results and carry out goal descriptions well. On the other hand, both count on inference optimization (CLIP alignment and DDIM inversion), which takes more than 200 seconds (\textit{vs.} ours 9.2 seconds) for each editing task.

\vspace{-1ex}
\begin{table}[h]
\centering \tablestyle{1.2pt}{1.1}
    \begin{tabular}{cccccccccccccccccc}
        \toprule
        \textbf{Method} & ~ & \multicolumn{3}{c}{\textbf{EVR}} & ~ & \multicolumn{3}{c}{\textbf{GIER}} & ~ & \multicolumn{3}{c}{\textbf{MA5k}} & ~ & \multicolumn{4}{c}{\textbf{MagicBrush}} \\
        \cmidrule{3-5} \cmidrule{7-9} \cmidrule{11-13} \cmidrule{15-18} ~ & ~ & L1$\downarrow$ & DINO$\uparrow$ & CVS$\uparrow$ & ~ & L1$\downarrow$ & SSIM$\uparrow$ & CVS$\uparrow$ & ~ & L1$\downarrow$ & SSIM$\uparrow$ & LPIPS$\downarrow$ & ~ & L1$\downarrow$ & DINO$\uparrow$ & CVS$\uparrow$ & CTS$\uparrow$ \\
        \midrule
        Text2LIVE & ~ & \underline{0.169} & 66.19 & 78.22 & ~ & \textbf{0.126} & \underline{58.32} & 79.32 & ~ & \underline{0.165} & 57.62 & 0.342 & ~ & \textbf{0.071} & \textbf{83.35} & \underline{89.71} & 23.59 \\
        Null-Inv & ~ & 0.174 & \underline{69.24} & 78.35 & ~ & 0.149 & 58.24 & 82.33 & ~ & 0.179 & \underline{61.36} & \underline{0.335} & ~ & \underline{0.073} & 81.72 & 87.24 & 27.62 \\
        InsPix2Pix & ~ & 0.189 & 67.82 & \underline{81.38} & ~ & 0.144 & 57.51 & \underline{86.63} & ~ & 0.176 & 58.92 & 0.359 & ~ & 0.101 & 71.46 & 85.22 & \underline{29.34} \\
        MGIE & ~ & \textbf{0.163} & \textbf{71.49} & \textbf{81.73} & ~ & \underline{0.135} & \textbf{59.24} & \textbf{88.59} & ~ &  \textbf{0.133} & \textbf{66.25} & \textbf{0.298} & ~ & 0.082 & \underline{82.22} & \textbf{91.14} & \textbf{30.40} \\
        \bottomrule
    \end{tabular}
    \vspace{-1ex}
\end{table}

\paragraph{Evaluating Image Editing via FID.}
As ground-truth goal images are available, we also calculate the Fréchet inception distance (FID) for editing results under the zero-shot or fine-tuned evaluation. However, the differences are all pretty limited. Since most editing results still resemble the original input images, it is difficult for FID to determine their authenticity. These results indicate that FID is insufficient to compare the quality of image editing.

\vspace{-1ex}
\begin{table}[h]
\centering \tablestyle{1.2pt}{1.1}
    \begin{tabular}{ccccccccccc}
        \toprule
        \textbf{Method} & ~ & \multicolumn{4}{c}{\textbf{Zero-shot}} & ~ & \multicolumn{4}{c}{\textbf{Fine-tuned}} \\
        \cmidrule{3-6} \cmidrule{8-11} ~ & ~ & EVR & GIER & MA5k & MagicBrush & ~ & EVR & GIER & MA5k & MagicBrush \\
        \midrule
        InsPix2Pix & ~ & \textbf{6.19} & \textbf{5.61} & 5.91 & 5.69 & ~ & \textbf{5.31} & \textbf{5.31} & \textbf{5.30} & 5.64 \\
        LGIE & ~ & 6.67 & 5.69 & \underline{5.80} & \textbf{5.31} & ~ & \underline{5.32} & \underline{5.42} & 5.59 & \underline{5.48} \\
        MGIE & ~ & \underline{6.45} & \underline{5.64} & \textbf{5.48} & \underline{5.61} & ~ & 5.53 & 5.59 & \underline{5.41} & \textbf{5.42} \\ 
        \bottomrule
    \end{tabular}
    \vspace{-1ex}
\end{table}

\paragraph{Part-of-Speech Distribution.}
We investigate part-of-speech (POS) distributions\footnote{We adopt flairNLP (\url{https://github.com/flairNLP/flair}) as the part-of-speech tagger.} of input instructions and our derived expressive instructions. In general, input instructions involve more nouns but fewer adjectives. In contrast, our expressive instructions can portray concrete edited scenes in detail via more adjectives. The original instructions are also dominated by verbs, which are challenging to perceive. The derivation helps them to be more understandable as adverbs. Moreover, we effectively decrease the number of ambiguous pronouns. More than 68\% pronouns (only 13\% in our expressive instructions) are unresolvable in input instructions\footnote{We apply AllenNLP (\url{https://github.com/allenai/allennlp}) for coreference resolution.}, where the model can not have explicit goals.

\begin{figure}[h]
\centering
    \vspace{-1ex}
    \includegraphics[width=.66\linewidth]{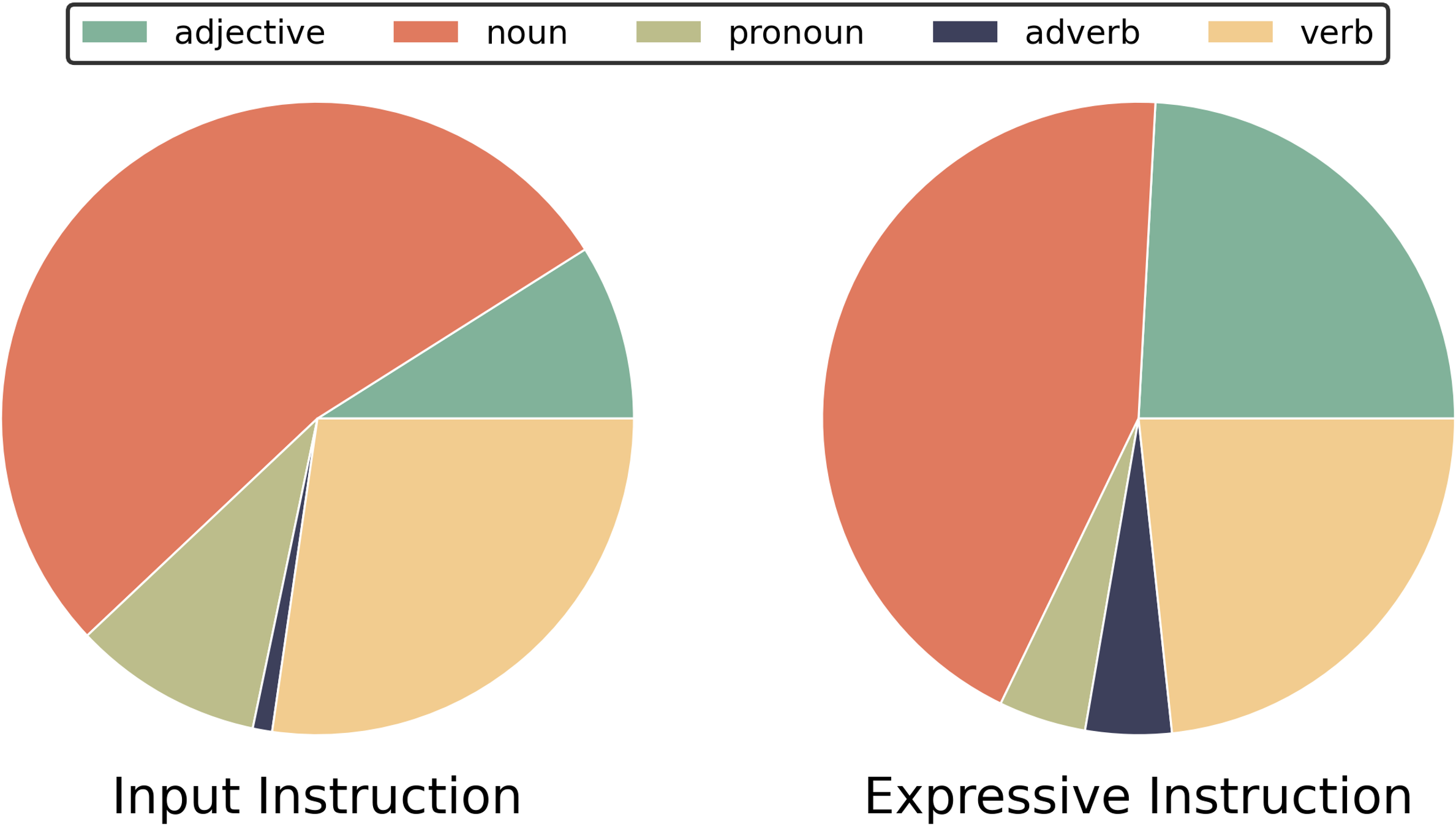}
    \vspace{-1ex}
\end{figure}

\paragraph{Unseen Editing Operation.}
Since there is no removal or photo optimization in IPr2Pr, InsPix2Pix has failed due to the shortage of training examples. Our MGIE is able to handle such editing via the visual-aware derivation of MLLM. We can accurately remove ``\textit{the boy in red shirt}'' or ``\textit{lighten out the yellow tone}'', which demonstrates better generalizability for unseen operations. More qualitative comparisons can be found on our project website\footref {foot:project}.

\clearpage

\begin{figure}[h]
\centering
    \includegraphics[width=.595\linewidth]{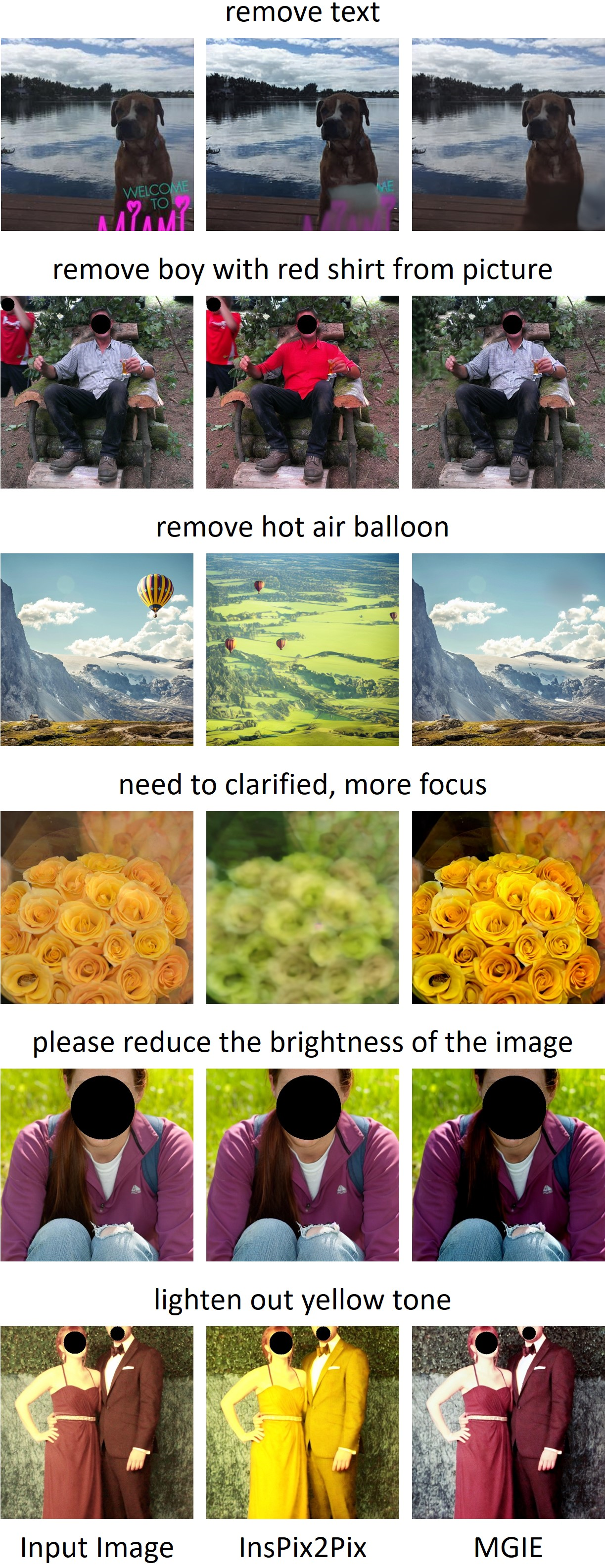}
\end{figure}

\clearpage

\paragraph{Ablation Study of Training Loss.}
There are two training losses, instruction loss ($\mathcal{L}_\text{ins}$) and editing loss ($\mathcal{L}_\text{edit}$), in our MGIE. $\mathcal{L}_\text{edit}$ is necessary for training to produce the editing result. Without $\mathcal{L}_\text{ins}$, it will derive full but lengthy guidance to lead $\mathcal{L}_\text{edit}$. However, both LGIE and MGIE drop significantly; LGIE even performs worse than the baseline. This underscores the prominence of learning concise expressive instructions, which offer succinct and relevant guidance. Besides, lengthy instructions via the MLLM will incur additional overhead (29.4 \textit{vs.} ours 9.2), resulting in an inefficient inference.

\vspace{-1ex}
\begin{table}[h]
\centering \tablestyle{1.2pt}{1.1}
    \begin{tabular}{ccccccccc}
        \toprule
        \textbf{Method} & \textbf{Setting} & ~ & \multicolumn{2}{c}{\textbf{MA5k}} & ~ & \multicolumn{3}{c}{\textbf{MagicBrush}} \\
        \cmidrule{4-5} \cmidrule{7-9} ~ & ~ & ~ & SSIM$\uparrow$ & LPIPS$\downarrow$ & ~ & DINO$\uparrow$ & CVS$\uparrow$ & CTS$\uparrow$ \\ 
        \midrule
        \multicolumn{2}{c}{InsPix2Pix} & ~ & 58.92 & 0.359 & ~ & 71.46 & 85.22 & 29.34 \\
        \midrule
        \multirow{2}{*}{LGIE} & - $\mathcal{L}_\text{ins}$ & ~ & 57.59 & 0.386 & ~ & 70.79 & 83.21 & 28.66 \\
        ~ & + $\mathcal{L}_\text{ins}$ & ~ & \textbf{64.60} & \textbf{0.327} & ~ & \textbf{80.90} & \textbf{88.87} & \textbf{30.10} \\
        \midrule
        \multirow{2}{*}{MGIE} & - $\mathcal{L}_\text{ins}$ & ~ & 58.18 & 0.365 & ~ & 71.50 & 85.19 & 29.11  \\
        ~ & + $\mathcal{L}_\text{ins}$ & ~ & \textbf{66.25} & \textbf{0.298} & ~ & \textbf{82.22} & \textbf{91.14} & \textbf{30.40} \\
        \bottomrule
    \end{tabular}
    \vspace{-1ex}
\end{table}

\paragraph{Adding New Object.}
MGIE also supports adding new objects that are not present in the input and placing them in reasonable positions. For instance, the ``\textit{hat}'' is put on the girl's head, and the ``\textit{river}'' is added along with the grass. More surprisingly, the appended ``\textit{fireworks}'' further makes the beach colorful, which drives the night scene coherent and visually appealing.

\begin{figure}[h]
\centering
    \vspace{-1ex}
    \includegraphics[width=.8165\linewidth]{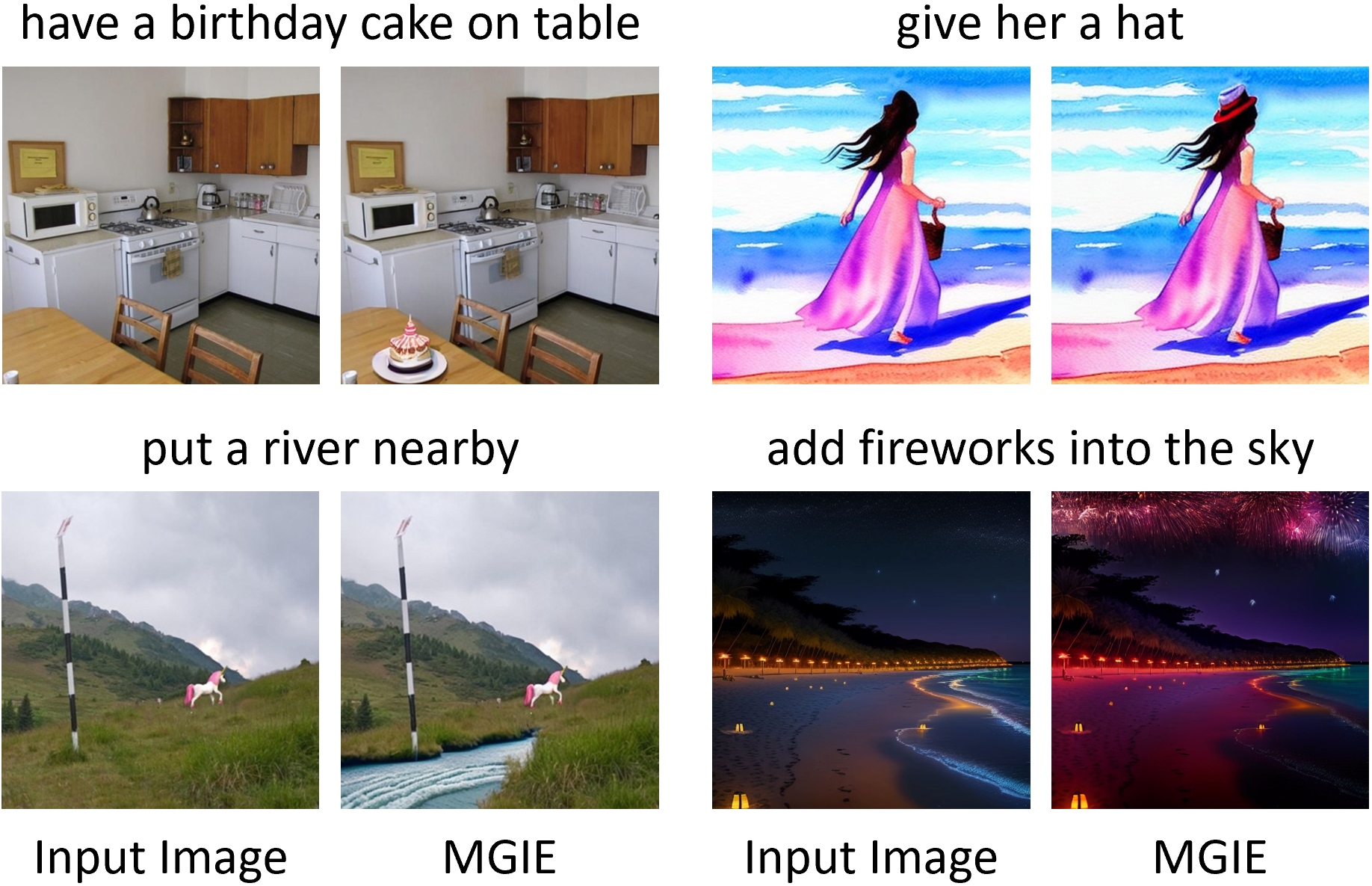}
    \vspace{-1ex}
\end{figure}

\clearpage

\paragraph{Transferring Image Texture/Color/Emotion.}
We attempt transferring visual patterns of images, also controlled through human instructions. For texture, we follow CLVA~\citep{fu2022ldast} and adopt the style prompt ``\textit{make the whole image as texture}~\texttt{[ins]}''. InsPix2Pix can only do limited transfer, but MGIE shows clear visual attributes (\textit{e.g.,} ``\textit{orange}'' or ``\textit{pinkish}'') as well as the complex ``\textit{colorful circular round}''. We perform fine-grained color manipulation, including ``\textit{glasses frame}'' or ``\textit{hair}''. However, the baseline even alters the whole color. For global colorization~\citep{chang2023l-cad}, both InsPix2Pix and our MGIE cannot present appealing results, which indicates the need for fine-tuning. Transferring the emotion is more challenging as the model has to perceive the latent semantics. We are able to illustrate the visual concept of ``\textit{bright day}'' or ``\textit{chaotic and confused}'' as the beach in the early morning or the gloomy street at night. MGIE can also transform from the cozy snowy day into suspenseful and thrilling through ``\textit{nightmare and scared}''. Although exhibiting promising potential, it still requests more profound texture/emotion perception for each specific goal. We leave them as future research for creative visual editing~\citep{weng2023aif}.

\begin{figure}[h]
\centering
    \vspace{-1ex}
    \includegraphics[width=.595\linewidth]{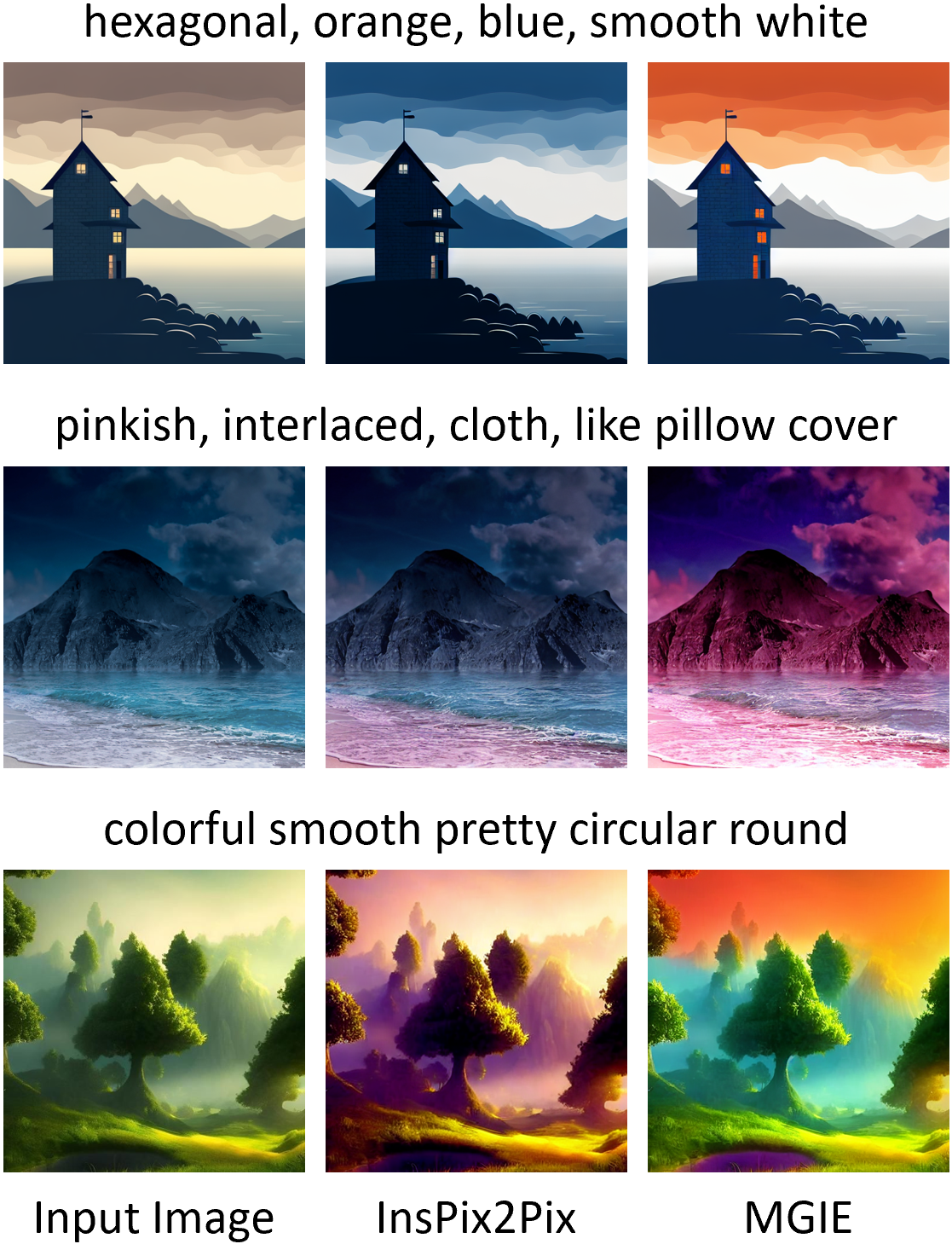}
    \vspace{-1ex}
\end{figure}

\noindent \centerline{\textit{color/emotion results on the next page}}

\clearpage

\begin{figure}[h]
\centering
    \includegraphics[width=.595\linewidth]{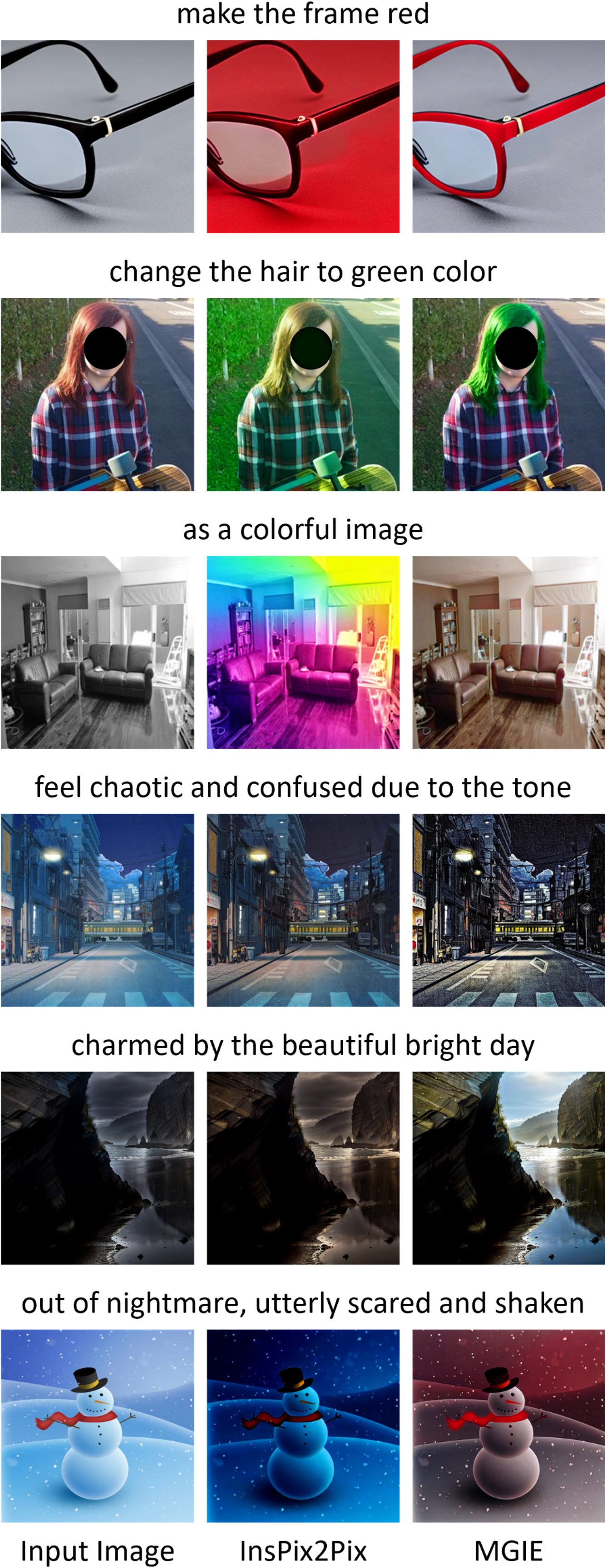}
\end{figure}

\clearpage

\section{Detailed Experimental Setup}
\paragraph{Edit Head to Joint the MLLM and the Diffusion Model.}
These appended visual tokens \texttt{[IMG]} are treated as the latent imagination of the editing goal from the MLLM but in the language modality. Inspired by GILL~\citep{koh2023gill}, we consider an edit head $\mathcal{T}$ to transform them into actual visual guidance. $\mathcal{T}$ is a lightweight 4-layer Transformer, which takes word embeddings $e$ and hidden states $h$ of \texttt{[IMG]} as the input and generates the visual imagination $\{u_1, ..., u_L\}$, conditioned on learnable query embeddings $\{q_1, ..., q_L\}$. As our diffusion model is inherited from StableDiffusion~\citep{rombach2022sd}, we apply the same $L=77$, and the dimension of $u$ is 768.

\begin{figure}[h]
\centering
    \vspace{-1ex}
    \includegraphics[width=.3335\linewidth]{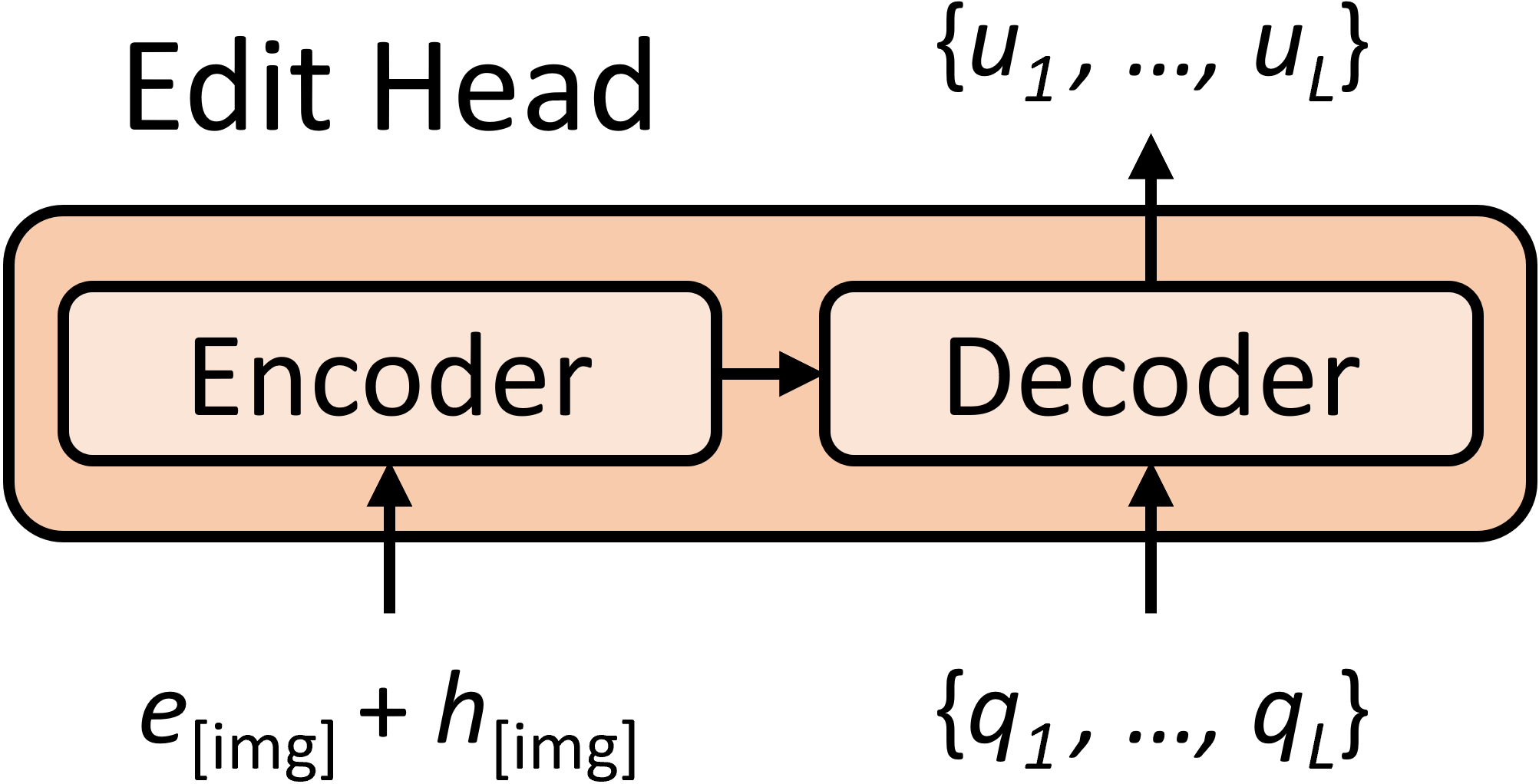}
    \vspace{-1ex}
\end{figure}

\paragraph{Editing Loss of the Diffusion Model.}
Our diffusion model is built upon latent diffusion $\mathcal{F}$~\citep{rombach2022sd}, which operates the latent space of the variational autoencoder (VAE). For the goal image $\mathcal{O}$, the diffusion process keeps adding noises to the encoded $o=\text{Enc}_\text{VAE}(\mathcal{O})$ and produces a noisy latent $z_t$. Our target is to learn the UNet $\epsilon_\theta$ that predicts the added noise according to the input image $v=\text{Enc}_\text{VAE}(\mathcal{V})$ and the visual imagination $\{u\}$ from the MLLM. The learning objective is:
\begin{equation} \notag
    \mathcal{L}_\text{edit} = \mathbb{E}_{o, v, \{u\}, \epsilon \sim \mathcal{N}(0, 1), t} \left[ ||\epsilon - \epsilon_\theta(z_t, t, v, \{u\})||^2_2 \right].
\end{equation}
Following InsPix2Pix~\citep{brooks2023ins-pix2pix}, we leverage the classifier-free guidance~\citep{ho2021classifier-free}, which combines both conditional and unconditional (a fixed null value $\varnothing$) denoising. During inference, we let the score estimation $s_\theta$ extrapolate toward the conditional yet keep away from the unconditional guidance. Since there are two conditionings ($v$ for image and $\{u\}$ for instruction), our modified $s_\theta$ should be:
\begin{equation}
\begin{split} \notag
    s_\theta(z_t, v, \{u\}) &= s_\theta(z_t, \varnothing, \varnothing) \\
    &+ \alpha_\mathcal{V} \cdot (s_\theta(z_t, v, \varnothing) - s_\theta(z_t, \varnothing, \varnothing)) \\
    &+ \alpha_\mathcal{X} \cdot (s_\theta(z_t, v, \{u\}) - s_\theta(z_t, v, \varnothing)),
\end{split}
\end{equation}
where we randomly set $v=\varnothing$, $\{u\}=\varnothing$, or both $=\varnothing$ for 5\% of data during training. $\alpha_\mathcal{V}$ and $\alpha_\mathcal{X}$ are guidance scales to control the trade-off between input image similarity and instruction alignment. By default, we use $\alpha_\mathcal{V}=1.5$ and $\alpha_\mathcal{X}=7.5$.

\paragraph{Training Cost.}
Our MGIE training requires 26 epochs to converge, and InsPix2Pix has 20 epochs (from their released checkpoint). Both MGIE and InsPix2Pix take a similar 1.6 hours per epoch on our node (8 NVIDIA A100 GPUs), where the overall training can be done in two days.

\paragraph{Human Evaluation.}
We sample 100 examples (25 for each dataset) to conduct our human evaluation. Each task is assigned 3 annotators, who rank across baselines and our MGIE, to avoid potential bias. We require workers to have a 97\% approval rate and over 500 approved tasks to ensure quality. The worker is awarded \$5 for each task (5 examples) and takes 21 minutes on average to complete.

\section{Ethics Discussion and Limitation}
In this paper, we leverage multimodal large language models (MLLMs) with the diffusion model to enhance instruction-based image editing. Even though our work benefits creative visual applications, there are still limitations that should be taken into consideration when interpreting the results. Since our MGIE is built upon pre-trained foundation models, it is possible to inherit bias from LLaVA and StableDiffusion. To mitigate this issue, we make the derived expressive instruction concise through summarization and update the MLLM together with the diffusion model. This end-to-end learning can also reduce the potential harmfulness since the hallucination from the LM will not be expressed over the editing. We can incorporate the safety checker~\citep{rombach2022sd} to filter out offensive results during post-processing as the final line of defense. From the perspective of editing, there are some challenging cases. Compositional command is hard to accomplish in a single step. Our MGIE can successfully remove the left sign but not the subsequent manipulation. In addition, the ability of language grounding (\textit{e.g.,} only the potato should be replaced), as well as numerical perception (\textit{e.g.,} just add to one cupcake), can be improved for more accurate targeting. We leave these directions as future research to achieve more practical and powerful instruction-based image editing.

\begin{figure}[h]
\centering
    \vspace{-1ex}
    \includegraphics[width=.914\linewidth]{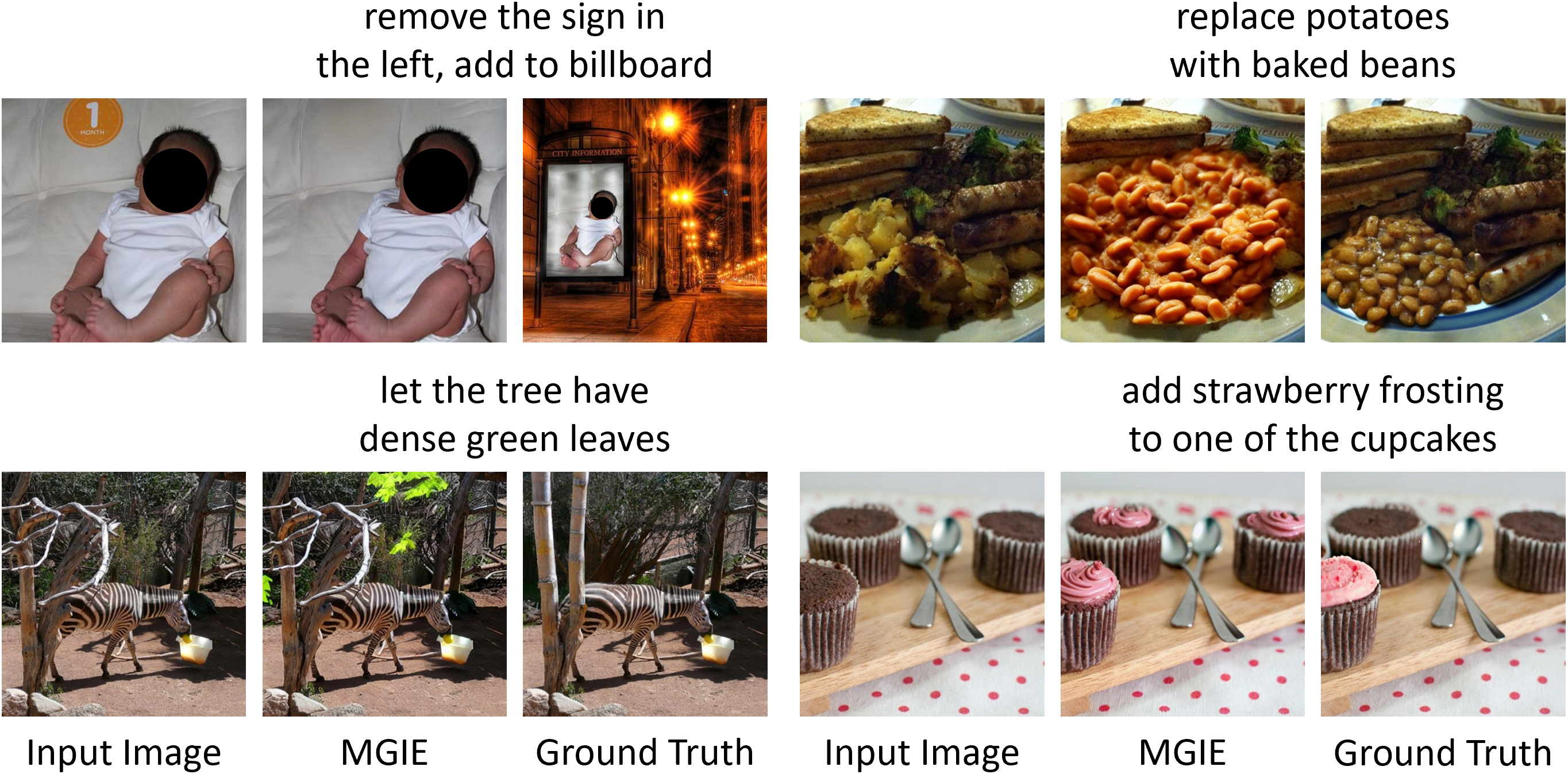}
    \vspace{-1ex}
\end{figure}

\end{document}